\documentclass[sigconf]{acmart}
\AtBeginDocument{%
  }

\usepackage{times}
\usepackage{latexsym}

\usepackage[T1]{fontenc}

\usepackage[utf8]{inputenc}

\usepackage{microtype}

\usepackage{graphicx}
\usepackage{xcolor}

\usepackage{amsmath}
\usepackage{amsthm}

\usepackage{booktabs}
\usepackage{multirow}
\usepackage{colortbl}

\usepackage{algorithm}
\usepackage{algpseudocode}

\usepackage{listings}

\usepackage{wrapfig}

\usepackage{tocloft}
\usepackage{pifont}
\lstdefinelanguage{json}{
  basicstyle=\ttfamily\small,
  numbers=left,
  numberstyle=\tiny\color{gray},
  stepnumber=1,
  numbersep=5pt,
  showstringspaces=false,
  breaklines=true,
  frame=single,
  stringstyle=\color{black},
  literate=
   *{0}{{{\color{blue}0}}}{1}
    {1}{{{\color{blue}1}}}{1}
    {2}{{{\color{blue}2}}}{1}
    {3}{{{\color{blue}3}}}{1}
    {4}{{{\color{blue}4}}}{1}
    {5}{{{\color{blue}5}}}{1}
    {6}{{{\color{blue}6}}}{1}
    {7}{{{\color{blue}7}}}{1}
    {8}{{{\color{blue}8}}}{1}
    {9}{{{\color{blue}9}}}{1}
    {:}{{{\color{red}{:}}}}{1}
    {,}{{{\color{red}{,}}}}{1}
    {"}{{{\color{brown}{"}}}}{1},
}

\usepackage[most]{tcolorbox}
\usepackage[table]{xcolor}

\newtcolorbox{myquote}{
  colback=gray!10,
  colframe=gray!80,
  boxrule=1.5pt,
  arc=2mm,
  left=5pt, right=5pt, top=5pt, bottom=5pt
}

\theoremstyle{definition}

\theoremstyle{remark}

\setcopyright{acmlicensed}
\copyrightyear{2018}
\acmYear{2018}
\acmDOI{XXXXXXX.XXXXXXX}
\acmConference[Conference acronym 'XX]{Make sure to enter the correct
  conference title from your rights confirmation email}{June 03--05,
  2018}{Woodstock, NY}
\acmISBN{978-1-4503-XXXX-X/2018/06}

\begin{document}

\title{Towards Faithful Reasoning in Comics for Small MLLMs}

\renewcommand\footnotetextcopyrightpermission[1]{}
\settopmatter{printacmref=false} 

\author{Chengcheng Feng}
\affiliation{%
  \institution{Duke Kunshan University}
  \country{China}
}

\author{Haojie Yin}
\affiliation{%
  \institution{Duke Kunshan University}
  \country{China}
}

\author{Yucheng Jin}
\affiliation{%
  \institution{Duke Kunshan University}
  \country{China}
}

\author{Kaizhu Huang}
\affiliation{%
  \institution{Duke Kunshan University}
  \country{China}
}
\renewcommand{\shortauthors}{Trovato et al.}

\begin{abstract}
Comic understanding presents a significant challenge for Multimodal Large Language Models (MLLMs), as the intended meaning of a comic often emerges from the joint interpretation of visual, textual, and social cues. This naturally motivates Chain-of-Thought (CoT) prompting, since explicit intermediate reasoning appears promising for integrating such heterogeneous signals. However, existing CoT methods are poorly matched to this structure: they tend to force interpretation into a single reasoning path before multiple cues have been jointly considered, often degrading performance, especially for small MLLMs. Our key idea is to explicitly preserve multi-cue interpretation during supervision construction, rather than collapsing comic understanding into a single reasoning chain.
To this end, we propose a two-stage framework for faithful comic reasoning in small MLLMs. First, we introduce \textbf{MoCoT}, a modular supervision construction framework that preserves multi-cue interpretation and turns it into more faithful supervision. Second, we propose \textbf{VERA}, a structured reward mechanism that turns such supervision into faithful reasoning behavior by aligning optimization with both reasoning faithfulness and answer correctness.
Extensive experiments on five benchmarks spanning comic understanding and broader humor-centric and abstract visual reasoning tasks demonstrate that our framework achieves strong results in the \(\leq\)4B regime, surpasses several 7B baselines, improves four small MLLMs by an average of \textbf{12.1\%} as a plug-in, and consistently enhances reasoning faithfulness while preserving inference efficiency.
Code is provided in the supplementary material.
\end{abstract}


\begin{teaserfigure}
  \includegraphics[width=1\linewidth]{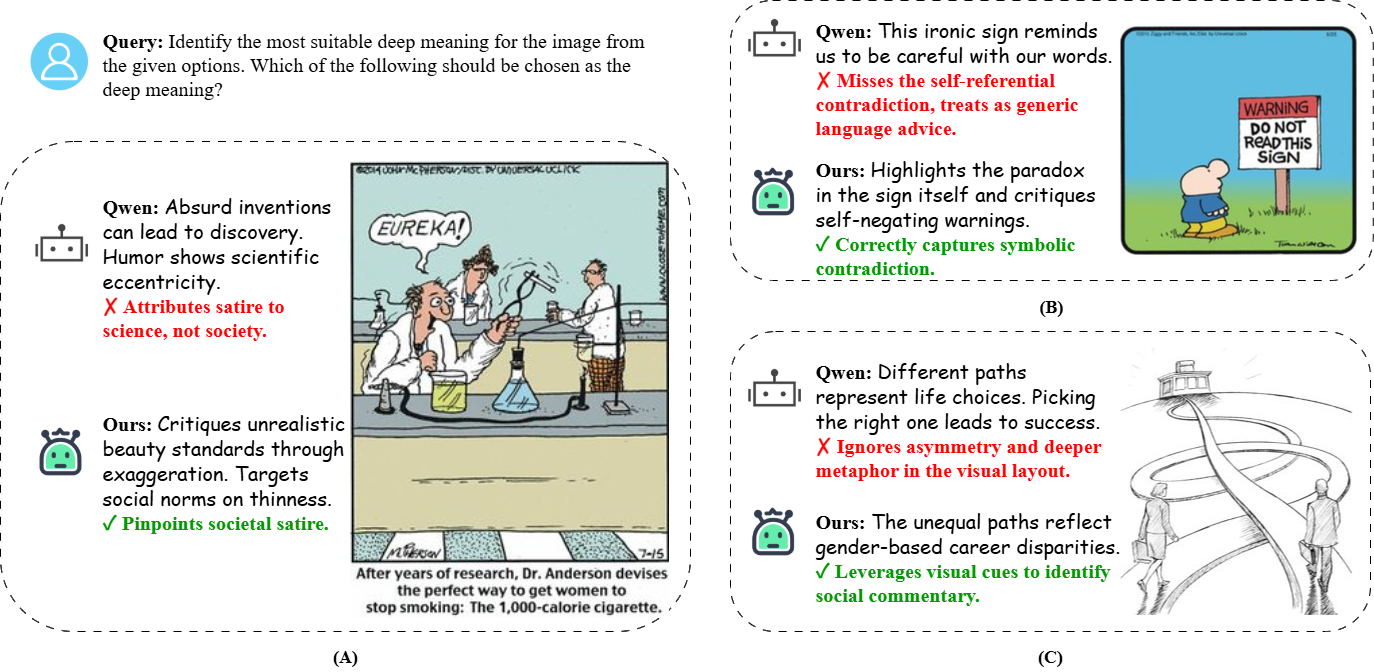}
  \caption{Comparison of reasoning failures in Qwen2.5-VL-3B (shown as Qwen in the figure) and our method. The examples highlight three typical error types under naive CoT prompting: (A) satirical target confusion, (B) symbolic misalignment, and (C) salient cue omission. In each case, Qwen's flawed reasoning leads to an incorrect answer, whereas our method produces more faithful reasoning and reaches the correct answer.}
  \label{quan}
\end{teaserfigure}




\maketitle

\section{Introduction}

\begin{figure*}[t]
  \centering
  \includegraphics[width=\textwidth]{pic/intro4.pdf}
  \caption{(A) Performance change from w/o CoT to w/ CoT on CII-Bench. In most cases, adding naive CoT degrades accuracy rather than improving it, with smaller MLLMs showing larger drops and greater instability. The complete numerical results are provided in Appendix~\ref{app:cii-bench-full}. (B) Our plug-in consistently improves accuracy across small MLLMs on DeepEval, compared with both w/ CoT and w/o CoT baselines.}
  \label{moti}
\end{figure*}

Comics constitute a distinctive multimodal narrative medium that compactly integrates visual scenes, textual content, and broader social context, so that meaning emerges through multiple interdependent cognitive operations rather than from a single flat stream of interpretation~\cite{cohn2020your,samson2008cognitive}. The process of comic understanding is illustrated by the comic in Figure~\ref{quan}(A). Its laboratory setting, white coats, scientific equipment, and the exclamation ``Eureka!'' initially suggest a scientific breakthrough, yet the caption about a ``1,000-calorie cigarette'' introduces an incongruity that cannot be resolved through literal reading alone. Correct interpretation requires jointly integrating visual cues, textual content, and social knowledge about body image and gender stereotypes, ultimately revealing the comic's satirical intent. More broadly, this dependence on cross-cue integration makes comic understanding a particularly challenging testbed for multimodal reasoning—one that current multimodal large language models (MLLMs) still struggle to handle~\cite{hu2024cracking,yang2024can,zhang-etal-2025-mllms,liu2024ii}.

This makes Chain-of-Thought (CoT) prompting~\citep{wei2022chain} a natural starting point: if comic understanding requires models to connect multiple interdependent cues into an implicit interpretation, then explicitly eliciting intermediate reasoning may seem promising for MLLMs. However, our experiments on the comic benchmark \textsc{CII-Bench}~\cite{zhang-etal-2025-mllms} reveal a counterintuitive result: \emph{naive CoT prompting often degrades performance, with small MLLMs suffering the most severe drop} (Figure~\ref{moti}(A)).


One possible explanation is that, while naive CoT makes reasoning explicit, it also encourages the model to commit too early to a single unfolding interpretation, even though comic understanding often requires keeping multiple cues and possible interpretations in play until they can be jointly resolved. In the example above, later information does not simply add to earlier observations, but changes how they should be interpreted. When this process is collapsed too early into one chain, omitted cues or inappropriate background assumptions can propagate through later steps and become difficult to correct. As a result, the model may overlook critical visual evidence, misread symbolic cues, or misidentify the satirical target (Figure~\ref{quan}), leading to an unfaithful reasoning process that ultimately yields incorrect answers. Together, these observations motivate us to ask \textbf{how faithful comic reasoning can be better supported, especially when interpretation depends on cross-cue integration}. A more precise definition of \textit{faithful comic reasoning} is provided in Section~\ref{sec:def}.

This challenge becomes especially pronounced in small models, where limited capacity makes faithful reasoning under cross-cue integration particularly difficult to sustain. Compared with larger models, small MLLMs not only perform worse on comic understanding~\cite{hu2024cracking,yang2024can}, but are also more vulnerable to degradation under naive CoT prompting. At the same time, they are the models most relevant to resource-constrained deployment. We therefore focus on small MLLMs in this work.





More concretely, supporting \emph{faithful comic reasoning} for small MLLMs requires resolving two distinct but coupled questions: \textsc{Q1}, \emph{faithful supervision construction}---how to construct supervision that is itself faithful to the multi-cue, context-sensitive nature of comic reasoning; and \textsc{Q2}, \emph{faithful reasoning learning}---how to use such supervision to induce reliable reasoning behavior in a small model. These two questions are related, but not identical.

For \textsc{Q1}, we introduce \textbf{MoCoT}, a modular reasoning construction framework designed to replace raw monolithic CoT supervision with a more structured alternative. MoCoT decomposes comic understanding into a small set of independent interpretive sub-problems, making explicit the implicit knowledge, background assumptions, and intermediate reasoning requirements entangled in the original question. It then organizes these heterogeneous reasoning components before forming the final rationale, so that the resulting supervision is a more faithful signal grounded in the multi-cue, context-sensitive nature of comic understanding.

For \textsc{Q2}, even when supervision is made more faithful, turning such supervision into reliable reasoning behavior remains fundamentally challenging. Supervised learning alone is insufficient, because during training the model learns to continue reasoning from ground-truth histories, whereas at inference time it must continue from its own generated reasoning. Once this self-generated reasoning begins to drift, faithfulness can quickly deteriorate~\cite{keneshloo2019deep}. By contrast, reinforcement learning (RL) optimizes the model directly on its own generated reasoning trajectories. However, naive RL still requires carefully designed reward signals; otherwise, optimization may fail to align with reasoning faithfulness.
For small MLLMs, there is an additional requirement: faithful reasoning should be learned without reproducing the full procedural form of supervision at inference time, since such procedural imitation can introduce unnecessary overhead and may hinder generalization~\cite{liu2024can}.

To address \textsc{Q2}, we do not require the model to reproduce the full modular reasoning process itself. Instead, we treat the final rationales produced by MoCoT as learning scaffolds. We then introduce \textbf{VERA}, a structured reward mechanism that explicitly encodes both reasoning faithfulness and answer correctness, so that optimization is guided not only by whether the prediction is correct, but also by whether the reasoning process remains coherent and well grounded. We instantiate this stage with GRPO~\cite{shao2024deepseekmath}, an RL algorithm, using VERA to reliably optimize the model on its own generated reasoning trajectories. As a result, VERA enables the model to internalize faithful reasoning patterns from MoCoT supervision and maintain them during generation, thereby addressing \textsc{Q2} while preserving inference efficiency.

Empirically, we validate our approach on five challenging benchmarks spanning comic understanding and broader humor-centric and abstract visual reasoning tasks, including meme understanding and editorial cartoon interpretation. On small MLLMs, our framework yields strong effectiveness gains: it achieves a new strong result in the \(\leq\)4B scale regime, and even surpasses several 7B baselines with and without CoT. As a plug-in, it further improves four small models by an average of 12.1\%, as shown in Figure~\ref{moti}(B). Beyond answer accuracy, the gains are also reflected in faithful reasoning itself: across all five benchmarks, our method consistently improves both reasoning faithfulness and final prediction performance over classical and state-of-the-art CoT baselines. Qualitative examples in Appendix~\ref{comqual} further show that our model better preserves the multi-cue and context-sensitive nature of comic understanding. These improvements do not come at the cost of efficiency, as our inference time remains comparable to vanilla CoT and substantially lower than prior methods that enhance CoT through heavier inference-time procedures. Together, these results show that structured supervision construction and process-level optimization provide a practical route toward more faithful, effective, and efficient small-model reasoning.

Our contributions are summarized as follows:

\begin{itemize}
    \item \textit{New insight and perspective.} We empirically reveal and analyze the structural mismatch between naive CoT-based reasoning and the multi-cue, context-sensitive nature of comic understanding, thereby formulating the task as one of faithful comic understanding.
    
    \item \textit{New supervision construction strategy.} We propose \textbf{MoCoT}, a modular reasoning construction strategy that preserves the structure of comic interpretation and turns it into more faithful supervision for comic understanding.
    
    \item \textit{New reasoning learning strategy.} We identify the challenge of reliably translating faithful supervision into faithful reasoning behavior in small MLLMs. To address this problem, we propose \textbf{VERA}, a structured reward mechanism that helps models internalize faithful reasoning patterns from MoCoT supervision and sustain them during generation.
    
    \item \textit{Compelling empirical results.} Extensive experiments on five benchmarks show that our framework consistently improves both reasoning faithfulness and prediction quality. It achieves a strong result in the \(\leq\)4B regime, surpasses several 7B baselines, and serves as an effective plug-in that improves four small MLLMs by an average of 12.1\% while preserving inference efficiency.
\end{itemize}

A discussion of related work is provided in Appendix~\ref{app:related}.

\section{Method}

\begin{figure*}[t]
  \centering
  \includegraphics[width=1\linewidth]{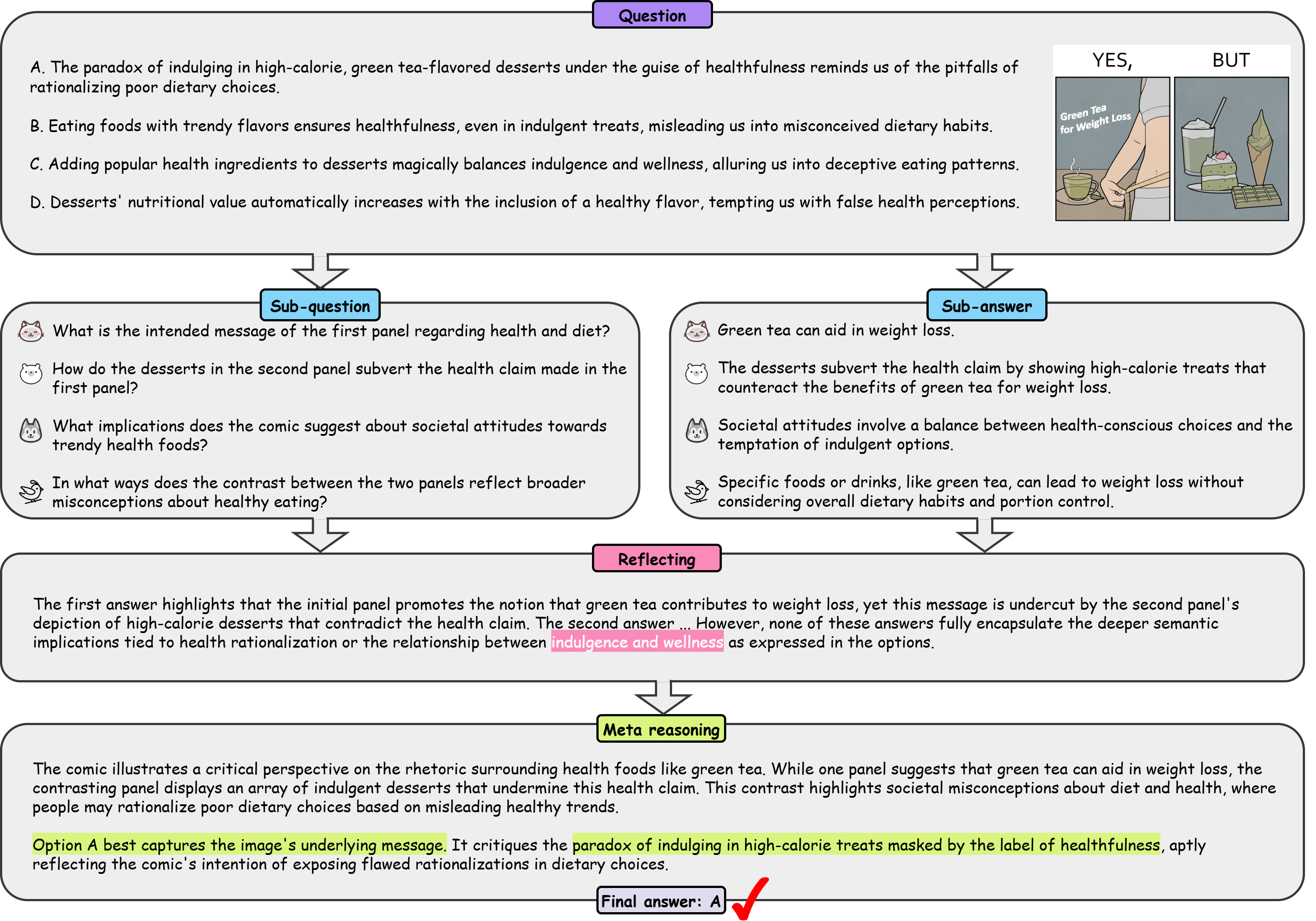}
  \caption{Our proposed MoCoT pipeline constructs faithful supervision for comic-based VQA by decomposing the original problem into complementary sub-questions, resolving them through localized image-grounded reasoning, and then performing reflective meta-reasoning with post-hoc consistency verification. A full case comparison between CoT and MoCoT is provided in Appendix~\ref{app:case}.}
  \label{mocot}
\end{figure*}


\subsection{Defining Faithful Comic Reasoning}
\label{sec:def}

This subsection defines what we mean by \emph{faithful comic reasoning}, which serves as the central principle of our framework. Comic meaning often depends on the joint resolution of visual cues, textual content, and social knowledge. What is therefore needed is a reasoning process that properly preserves and integrates these complementary interpretive elements, so that the model can reach the correct answer through a more reliable and better-grounded understanding of the comic. We therefore treat faithful reasoning, rather than answer accuracy alone, as the core objective underlying both \textsc{Q1} and \textsc{Q2}.

In this work, we define \emph{faithful comic reasoning} as reasoning that satisfies three properties. First, it should exhibit \emph{evidence grounding}, remaining anchored in the visual and textual cues that are genuinely relevant to the comic rather than introducing unsupported or hallucinated claims. Second, it should preserve \emph{symbolic and narrative coherence}, integrating implicit background knowledge and social context in a way that is compatible with the comic's non-literal, humorous, or satirical intent. Third, it should maintain \emph{procedural consistency}, so that the final answer is supported by the reasoning process that produces it, rather than justified retrospectively by a plausible-looking explanation. Faithfulness in this sense concerns the validity of the reasoning process itself, rather than the length or verbosity of the generated rationale.

This definition provides the criterion for the two components of our framework: MoCoT targets faithful supervision construction for \textsc{Q1}, while VERA targets faithful reasoning learning for \textsc{Q2}.

\subsection{MoCoT: Faithful Supervision Construction}
\label{sec:mocot}

This subsection addresses \textsc{Q1}: how to construct supervision that remains faithful to the multi-cue and context-sensitive nature of comic understanding. Our key idea is that faithful supervision for comics should not be generated as a single direct rationale from the outset. Instead, it should preserve the intermediate interpretive judgments required for understanding the comic before they are integrated into a final explanation. To this end, we propose a structured supervision construction framework MoCoT, as illustrated in Figure~\ref{mocot}. Given a comic-based Visual Question Answering (VQA) instance $\mathcal{X}=(I,Q)$, where $I$ is the comic image and $Q$ is the associated question, MoCoT constructs a final rationale $R$ through four steps, described below.

\textit{Step 1: Subgoal planning via question decomposition.}
MoCoT first analyzes the original question in the context of the comic image and decomposes it into a small set of complementary sub-questions, denoted as \(\mathcal{Q}_{\mathrm{sub}}=\{q_k\}_{k=1}^{K}\), where \(\mathcal{Q}_{\mathrm{sub}}\) denotes the set of sub-questions, \(q_k\) is the \(k\)-th sub-question, and \(K\) is the total number of sub-questions.Intuitively, this step serves as a planning stage: before attempting the final answer, the model identifies which intermediate interpretive checkpoints must be resolved in order to answer the original question reliably. The goal is therefore not to split the problem by modality or by answer option. Instead, each sub-question makes explicit one intermediate interpretive judgment required for solving the comic, such as establishing a key visual fact, clarifying a symbolic or narrative implication, or surfacing the background assumption behind the humor or satire. In this way, Step~1 converts one under-specified question into several more explicit intermediate reasoning targets. This is crucial for faithfulness because it prevents the model from jumping too early to a single answer-oriented interpretation. Instead, multiple partial interpretive needs are kept explicit and can be resolved separately before they are integrated into the final rationale.

\textit{Step 2: Localized execution through image-grounded CoT.}
Each sub-question is then answered independently through a localized chain-of-thought process, denoted as \((r_k,a_k)=\mathcal{E}_k(I,q_k)\), where \(\mathcal{E}_k\) denotes the execution module applied to sub-question \(q_k\), \(r_k\) is the resulting rationale for that sub-question, and \(a_k\) is the corresponding provisional answer. Importantly, this stage is not a direct jump from the sub-question to a conclusion. Instead, the model is prompted to first identify the relevant visual elements in the image, then reason step by step, and finally produce the sub-answer. This design encourages intermediate reasoning to emerge from localized evidence rather than from immediate answer-oriented speculation. Because each execution step is centered on a decomposed sub-question, the reasoning scope is narrower than that of the original problem, which helps reduce omitted cues, unsupported inferential jumps, and loosely connected interpretations. In this way, Step~2 primarily strengthens \emph{evidence grounding} at the level of intermediate supervision.

\begin{figure*}[t]
  \centering
  \includegraphics[width=1\linewidth]{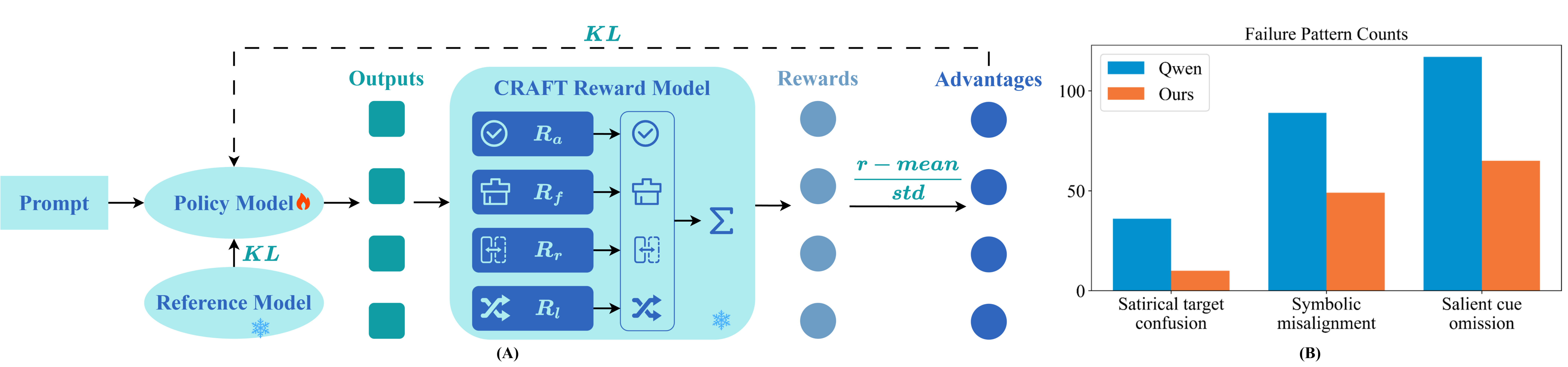}
  \caption{(A) VERA-guided GRPO optimization loop. Given an input prompt, the policy model samples multiple candidate outputs, which are scored by VERA. The resulting rewards are normalized into group-relative advantages, and KL regularization with a reference model stabilizes policy updates. (B) Reduction of representative failure patterns under our framework.}
  \label{grpo}
\end{figure*}

\textit{Step 3: Reflective meta-reasoning for final inference.}
This step is necessary because the sub-answers are not equally reliable: some may be accurate and useful, while others may be incomplete, locally plausible but globally misleading, or uneven in importance. MoCoT therefore does not directly aggregate the intermediate sub-results into a final rationale. Instead, it first performs a reflective meta-reasoning stage that diagnoses the quality of the sub-answers before final inference. Based on this diagnosis, a meta-reasoning module produces a pre-verification final rationale, denoted as \(\tilde{R}=\mathcal{M}\bigl(I,Q,\{(r_k,a_k)\}_{k=1}^{K}\bigr)\), where \(\mathcal{M}\) denotes the meta-reasoning module and \(\tilde{R}\) denotes the pre-verification final rationale. This design encourages a form of critical reflection: before committing to a final answer, the model must revisit and judge the quality of its intermediate reasoning. This is particularly important for comics, where the intended meaning often arises not from any single local clue, but from how multiple partial interpretations constrain, revise, and complement one another. By separating diagnosis from final inference, MoCoT reduces the risk that weak intermediate reasoning is propagated uncritically into the final rationale and supports a more reliable global interpretation than direct answer generation alone. Step~3 primarily improves \emph{symbolic and narrative coherence} at the global level.

\textit{Step 4: Post-hoc consistency verification.}
Finally, MoCoT applies a post-hoc verification stage that operates only on the final textual rationale and the predicted answer, without revisiting the image. Specifically, the final rationale after verification is defined as \(R=\mathcal{V}(\tilde{R},A)\), where \(A\) is the predicted final answer, \(\mathcal{V}\) denotes the textual verification module, and \(R\) is the final rationale after verification. Stage~4 is designed specifically to strengthen \emph{procedural consistency}: it reduces answer--rationale mismatch and discourages post-hoc justifications that are textually inconsistent with the final prediction.

Taken together, MoCoT turns faithful supervision construction from direct rationale generation into a staged process of planning, grounding, reflection, and consistency control.

\subsection{VERA: Faithful Reasoning Learning}
\label{sec:vera}

This subsection addresses \textsc{Q2}: how to translate faithful supervision into reliable reasoning behavior in small MLLMs. While MoCoT provides more faithful supervision, such supervision does not automatically become faithful reasoning at inference time. The core difficulty is that MoCoT rationales act as training scaffolds, whereas the final model must generate its own reasoning under self-produced trajectories. We therefore introduce VERA, a structured reward mechanism that encourages the model to internalize the criteria of faithful reasoning rather than merely imitate the procedural traces of MoCoT supervision.

For a generated output \(o\), the structured reward is defined as \(R(o)=\lambda_1 R_{\mathrm{f}}(o)+\lambda_2 R_{\mathrm{a}}(o)+\lambda_3 R_{\mathrm{r}}(o)+\lambda_4 R_{\mathrm{l}}(o)\), where \(\lambda_i\) is non-negative weight satisfying \(\sum_{i=1}^{4}\lambda_i=1\), and each term captures a complementary aspect of reasoning quality. Specifically, $R_{\mathrm{f}}$ is a format correctness reward that ensures the output remains in a structured and evaluable form. $R_{\mathrm{a}}$ is an answer accuracy reward that preserves outcome validity by checking whether the final prediction matches the ground truth. $R_{\mathrm{r}}$ is a reasoning similarity reward that encourages the generated rationale to remain close to the faithful rationale patterns constructed by MoCoT. Finally, $R_{\mathrm{l}}$ is a logic consistency reward that encourages agreement between the selected answer and the generated inference rationale. To avoid rewarding plausible-sounding justifications for incorrect predictions, $R_{\mathrm{r}}$ is activated only when the final answer is correct.

Among these terms, $R_{\mathrm{r}}$ and $R_{\mathrm{l}}$ provide the main faithfulness signal. $R_{\mathrm{r}}$ encourages the model to stay close to the faithful reasoning patterns encoded in MoCoT supervision, thereby discouraging arbitrary or post-hoc rationales, while $R_{\mathrm{l}}$ strengthens \emph{procedural consistency} by requiring the generated rationale to support the selected answer. By contrast, $R_{\mathrm{f}}$ and $R_{\mathrm{a}}$ play supporting roles by keeping outputs structured and anchoring optimization to the correct task outcome.

Taken together, VERA turns faithful reasoning learning from answer-only optimization into a structured process of aligning generated reasoning with observable faithfulness criteria.

\begin{table*}[t]
\centering
\resizebox{\linewidth}{!}
{%
\begin{tabular}{lccccccccccccccc}
\toprule
\multicolumn{3}{c}{\textbf{Model}} & \multicolumn{1}{c}{\textbf{DeepEval}} & \multicolumn{1}{c}{\textbf{YesBut}} & \multicolumn{10}{c}{\textbf{CII-Bench}} \\
\cmidrule(lr){1-3}\cmidrule(lr){6-15}
\textbf{Name} & \textbf{\#Params} & \textbf{CoT} &  &  & \textbf{Overall} & \textbf{Life} & \textbf{Art} & \textbf{Soc.} & \textbf{Pol.} & \textbf{Env.} & \textbf{CTC} & \textbf{Pos.} & \textbf{Neg.} & \textbf{Neu.} \\
\midrule
\multicolumn{15}{c}{\textit{\textbf{7B Scale MLLMs}}} \\
\midrule
\multirow{2}{*}{LLaVA-1.6~\citep{liu2023improved}} & \multirow{2}{*}{7B} & \ding{55} & 17.1 & 56.9 & 30.2 & 23.4 & 37.5 & 28.1 & 29.2 & 50.0 & 29.6 & 26.1 & 30.2 & 33.8 \\
                           &                     & \ding{51} & 29.7 & 54.9 & 29.0 & 21.7 & 34.6 & 30.3 & 29.2 & 44.4 & 28.2 & 25.6 & 30.2 & 30.8 \\
\addlinespace
\multirow{2}{*}{XComposer-2.5~\citep{zhang2024internlm}} & \multirow{2}{*}{7B} & \ding{55} & 34.2 & 50.2 & 32.6 & 26.8 & 36.8 & 35.7 & 25.0 & 42.6 & 31.1 & 31.6 & 35.5 & 30.5 \\
                               &                     & \ding{51} & 36.2 & 45.5 & 32.6 & 30.3 & 32.4 & 34.6 & 33.3 & 40.7 & 30.4 & 31.6 & 35.1 & 30.8 \\
\addlinespace
\multirow{2}{*}{Qwen2.5-VL~\citep{qwen2.5-VL}} & \multirow{2}{*}{7B} & \ding{55} & 58.3 & 68.8 & 48.1 & 41.1 & 52.2 & 51.4 & 58.3 & 53.7 & 47.4 & 47.9 & 47.2 & 49.3 \\
                            &                     & \ding{51} & 63.3 & 70.4 & 45.8 & 39.0 & 45.6 & 50.8 & 45.8 & 57.4 & 45.9 & 44.4 & 46.0 & 46.6 \\
\addlinespace
\multirow{2}{*}{InternVL3~\citep{zhu2025internvl3}}  & \multirow{2}{*}{8B} & \ding{55} & 70.9 & 65.6 & 50.7 & 45.9 & 48.5 & 57.8 & 45.8 & 51.9 & 51.9 & 46.6 & 52.5 & 52.6 \\
                            &                     & \ding{51} & 67.8 & 66.4 & 47.7 & 42.9 & 46.3 & 55.1 & 37.5 & 57.4 & 45.2 & 46.2 & 47.9 & 48.9 \\
\midrule
\multicolumn{15}{c}{\textit{\textbf{$\leq$4B Scale MLLMs}}} \\
\midrule
\multirow{2}{*}{Mono~\citep{luo2024mono}} & \multirow{2}{*}{2B} & \ding{55} & 14.1 & 48.2 & 22.5 & 17.8 & 22.8 & 21.1 & 29.2 & 27.8 & 28.9 & 23.1 & 21.1 & 23.3 \\
                               &                     & \ding{51} & 20.1 & 32.8 & 10.7 & 8.6 & 13.2 & 7.0 & 12.5 & 13.0 & 15.6 & 11.1 & 8.7 & 12.4 \\
\addlinespace
\multirow{2}{*}{Ovis2~\citep{lu2024ovis}} & \multirow{2}{*}{2B} & \ding{55} & 31.7 & 53.8 & 36.3 & 32.0 & 33.1 & 43.8 & 37.5 & 48.2 & 31.9 & 35.5 & 34.7 & 38.7 \\
                       &                     & \ding{51} & 32.2 & 50.6 & 26.8 & 22.1 & 28.7 & 35.1 & 37.5 & 35.2 & 16.3 & 23.1 & 28.3 & 28.6 \\
\addlinespace
\multirow{2}{*}{InternVL2.5~\citep{chen2024expanding}} & \multirow{2}{*}{2B} & \ding{55} & 45.7 & 45.5 & 33.6 & 27.3 & 36.8 & 37.3 & 41.7 & 40.7 & 31.9 & 30.8 & 34.7 & 35.0 \\
                             &                     & \ding{51} & 42.7 & 48.2 & 33.3 & \underline{33.3} & 37.5 & 35.7 & 29.2 & 29.6 & 28.2 & 32.5 & 30.6 & 36.8 \\
\addlinespace
\multirow{2}{*}{Qwen2.5-VL~\citep{qwen2.5-VL}} & \multirow{2}{*}{3B} & \ding{55} & \underline{55.8} & 55.7 & \underline{41.8} & 32.5 & \underline{39.0} & \underline{44.3} & \textbf{54.2} & \underline{53.7} & \textbf{50.4} & \underline{39.7} & \underline{41.5} & \underline{44.0} \\
                            &                     & \ding{51} & 48.7 & \underline{57.7} & 36.2 & 31.2 & 33.8 & 34.6 & 37.5 & 50.0 & 43.7 & 37.2 & 31.7 & 39.9 \\
\addlinespace
\multirow{2}{*}{Phi-3.5~\citep{abdin2024phi}} & \multirow{2}{*}{4B} & \ding{55} & 35.7 & 56.9 & 33.1 & 26.8 & \underline{39.0} & 32.4 & 45.8 & 44.4 & 31.9 & 26.5 & 37.4 & 34.6 \\
                         &                     & \ding{51} & 30.7 & 51.0 & 22.1 & 14.7 & 31.6 & 21.1 & 29.2 & 27.8 & 23.0 & 22.2 & 20.8 & 23.3 \\
\addlinespace
\multirow{2}{*}{Gemma-3~\citep{gemma_2025}} & \multirow{2}{*}{4B} & \ding{55} & 35.2 & 51.0 & 30.5 & 26.8 & 34.6 & 31.4 & 45.8 & 35.2 & 26.7 & 23.5 & 34.3 & 32.7 \\
                         &                     & \ding{51} & 46.2 & 47.0 & 32.6 & 29.0 & 37.5 & 31.9 & 25.0 & 40.7 & 32.6 & 26.9 & 32.1 & 38.0 \\
\addlinespace
\rowcolor{gray!20}
\textbf{Ours} & 3B & -- & \textbf{64.3(\textcolor{green!50!black}{\scriptsize +15.2\%})} & \textbf{62.9(\textcolor{green!50!black}{\scriptsize +9.0\%})} & \textbf{44.7(\textcolor{green!50!black}{\scriptsize +6.9\%})} & \textbf{35.9} & \textbf{44.1} & \textbf{49.2} & \underline{50.0} & \textbf{55.6} & \underline{48.9} & \textbf{41.0} & \textbf{44.9} & \textbf{47.7} \\
\bottomrule
\end{tabular}%
}
\caption[]{Overall accuracy (\%) of different MLLMs (with \ding{51} and without \ding{55} CoT) and our method across three benchmarks: DeepEval, YesBut, and CII-Bench\footnotemark (evaluated by domains and emotions). The best and second-best results among $\leq$4B models are highlighted in \textbf{bold} and \underline{underlined}, respectively. Our method is highlighted with a \colorbox{gray!20}{gray background}.}
\label{dabiao}
\end{table*}
\footnotetext{Soc.=Society, Pol.=Politics, Env.=Environment, CTC=Chinese Traditional Culture, Pos.=Positive, Neg.=Negative, Neu.=Neutral.}

\subsection{Optimization and Training Pipeline}
\label{sec:optimization}

This subsection describes how MoCoT and VERA are combined during training. MoCoT provides faithful rationale--answer pairs as the supervision source, VERA defines the reward criteria, and GRPO optimizes the model over self-generated reasoning trajectories.

We first construct the optimization set from MoCoT outputs. Specifically, for each training instance, we retain only those MoCoT-generated final rationale--answer pairs whose predicted answer matches the ground-truth answer. In other words, the optimization stage uses the final faithful rationales produced by MoCoT, rather than the full intermediate construction process. This design is important for two reasons. First, it keeps the learning target outcome-consistent, yielding a cleaner set of faithful rationale--answer pairs for optimization. Second, it avoids forcing the model to imitate the entire procedural form of MoCoT, which would introduce unnecessary overhead and may reduce generalization. These retained final rationale--answer pairs are then used as reference targets for VERA-guided optimization.

We then instantiate faithful reasoning learning with GRPO. The GRPO-based optimization loop with VERA rewards is illustrated in Figure~\ref{grpo}. For each input, the current policy model samples multiple candidate outputs, each containing a reasoning trace and a final answer. VERA is applied to score these outputs according to format correctness, answer accuracy, reasoning similarity, and logic consistency. GRPO estimates relative advantages by comparing the rewards of outputs generated for the same input, and updates the policy to increase the likelihood of higher-reward outputs while suppressing lower-reward ones. In this way, optimization encourages the model to internalize faithful comic reasoning patterns under self-generated trajectories. Pseudocode of our pipeline is provided in Appendix~\ref{app:pseudocode}.
\section{Experiments}

\subsection{Experiment Settings}

\textbf{Datasets and Benchmarks.}
We evaluate our framework on five benchmarks covering comic understanding, broader humor-centric and abstract visual reasoning, and diverse evaluation formats.

We use three comic-based multiple-choice benchmarks: DeepEval~\citep{yang2024can}, YesBut v2~\citep{liang2025yes} (abbreviated as YesBut), and CII-Bench~\citep{zhang-etal-2025-mllms}. DeepEval focuses on deep semantic inference, YesBut emphasizes contradiction and narrative reversal, and CII-Bench evaluates Chinese-language and culturally grounded comic understanding.

We further include two non-MCQ benchmarks to test transfer beyond standard comic QA. MemeCap~\citep{hwang2023memecap} evaluates open-ended meme interpretation, while NewYorker~\citep{hessel2023androids} evaluates discriminative humor understanding through caption--cartoon matching.

For MoCoT construction and GRPO fine-tuning, we randomly sample 80\% of DeepEval and YesBut and further split them into training and validation sets. The remaining portions of DeepEval and YesBut, together with CII-Bench, MemeCap, and NewYorker, are used exclusively for evaluation.

\begin{table*}[t]
\centering
\resizebox{\linewidth}{!}{
\begin{tabular}{l ccc ccc ccc ccc cccc}
\toprule
\multirow{2}{*}{\textbf{Method}} &
\multicolumn{3}{c}{\textbf{DeepEval}} &
\multicolumn{3}{c}{\textbf{YesBut}} &
\multicolumn{3}{c}{\textbf{CII-Bench}} &
\multicolumn{3}{c}{\textbf{NewYorker}} &
\multicolumn{4}{c}{\textbf{MemeCap}} \\
\cmidrule(lr){2-4}
\cmidrule(lr){5-7}
\cmidrule(lr){8-10}
\cmidrule(lr){11-13}
\cmidrule(lr){14-17}
&
\textbf{ACC}$\uparrow$ & \textbf{CCS}$\uparrow$ & \textbf{USR}$\downarrow$ &
\textbf{ACC}$\uparrow$ & \textbf{CCS}$\uparrow$ & \textbf{USR}$\downarrow$ &
\textbf{ACC}$\uparrow$ & \textbf{CCS}$\uparrow$ & \textbf{USR}$\downarrow$ &
\textbf{ACC}$\uparrow$ & \textbf{CCS}$\uparrow$ & \textbf{USR}$\downarrow$ &
\textbf{BLEU-4}$\uparrow$ & \textbf{ROUGE-L}$\uparrow$ & \textbf{BERT-F1}$\uparrow$ & \textbf{USR}$\downarrow$ \\

\midrule
CoT~\citep{wei2022chain} &
48.7 & 70.4 & 14.5 &
57.7 & 53.5 & 25.8 &
36.2 & 69.4 & 28.3 &
29.2 & 14.4 & 27.4 &
1.9 & 14.0 & 95.6 & 15.2 \\

ToT~\citep{yao2023tree} &
50.8 & \underline{79.0} & 22.4 &
59.3 & \underline{82.4} & 29.0 &
34.9 & 61.1 & 35.4 &
37.1 & \underline{23.4} & 30.6 &
4.5 & 20.3 & \textbf{97.4} & 15.3 \\

DDCoT~\citep{zheng2023ddcot} &
49.2 & 62.4 & 18.0 &
54.5 & 61.0 & 24.6 &
35.3 & 63.6 & 33.7 &
29.4 & 15.1 & 37.0 &
4.9 & 23.1 & 97.1 & 18.7 \\

CoT-SC~\citep{wangself} &
45.2 & 61.9 & 21.7 &
58.9 & 81.2 & 19.7 &
37.3 & 57.5 & 29.9 &
35.8 & 20.0 & 27.8 &
1.4 & 13.1 & 95.4 & 14.9 \\

GoT~\citep{besta2024graph} &
49.8 & 62.8 & 13.5 &
59.3 & 79.8 & 18.7 &
\underline{39.9} & \underline{72.1} & \textbf{24.3} &
\underline{40.2} & \underline{23.4} & 28.7 &
3.3 & 19.6 & 96.4 & 16.1 \\

CCoT~\citep{mitra2024compositional} &
51.3 & 60.4 & 19.6 &
59.7 & 58.9 & \underline{15.4} &
32.9 & 67.2 & 30.4 &
37.9 & 17.0 & 31.7 &
\underline{5.1} & 22.5 & \underline{97.2} & 19.3 \\

LAD~\citep{zhang2025let} &
42.7 & 15.4 & \underline{13.1} &
\underline{60.9} & 42.3 & 22.7 &
39.1 & 49.7 & 29.6 &
39.4 & 4.4 & \underline{26.8} &
2.0 & 13.8 & 95.6 & \underline{14.0} \\

Reasoner~\citep{zhang2025improve} &
\underline{52.8} & 74.9 & 16.8 &
60.1 & 78.6 & 16.1 &
38.2 & 67.9 & 27.6 &
38.6 & 21.1 & 27.8 &
4.8 & \underline{23.2} & \underline{97.2} & 14.9 \\

\rowcolor{gray!20}
\textbf{Ours} &
\textbf{64.3} & \textbf{85.9} & \textbf{12.9} &
\textbf{62.9} & \textbf{83.1} & \textbf{9.4} &
\textbf{44.7} & \textbf{72.5} & \underline{24.8} &
\textbf{41.1} & \textbf{24.1} & \textbf{26.6} &
\textbf{5.3} & \textbf{24.8} & \textbf{97.4} & \textbf{13.8} \\

\bottomrule
\end{tabular}
}
\caption{Comparison of different reasoning paradigms built upon Qwen2.5-VL-3B across five benchmarks.}
\label{tab:main_comparison}
\end{table*}

\noindent\textbf{Evaluation Metrics.}
For task performance, we use accuracy for multiple-choice benchmarks, and BLEU-4, ROUGE-L, and BERTScore F1 for open-ended generation.

We evaluate reasoning faithfulness along two automatically measurable aspects: evidence grounding and procedural consistency. Symbolic and narrative coherence, which is more difficult to capture reliably with automatic metrics, is assessed separately through qualitative analysis and human-annotated failure pattern statistics, as discussed in Section~\ref{fur} on Failure Patterns.

\textit{Unfaithful Statement Rate (USR).}
USR measures evidence grounding at the claim level. For each generated rationale, a multimodal judge extracts atomic claims and labels a claim as unsupported if it is not grounded in the image or the given question/options. USR is defined as the fraction of unsupported claims among all extracted claims.

\textit{Counterfactual Consistency Score (CCS).}
CCS measures procedural consistency at the trajectory level. Given a rationale and its predicted answer, a multimodal verifier checks whether the rationale entails the predicted option while not entailing a counterfactual option. A sample is counted as consistent if this condition holds. Pseudocode for USR and CCS are provided in Appendix~\ref{app:pseudocode}.


\noindent\textbf{Implementation Details.}\label{imple}
We run all experiments on 4 NVIDIA A800.
For MoCoT construction, we use Qwen2.5-VL-7B-Instruct~\citep{qwen2.5-VL} for Step 2 because this step is repeated for multiple sub-questions and accounts for most of the construction cost, while mainly requiring localized evidence-grounded reasoning. The remaining steps are implemented with gpt-4o-mini~\citep{hurst2024gpt}.

For GRPO fine-tuning, we adopt EasyR1~\citep{zheng2025easyr1} with Qwen2.5-VL-3B-Instruct~\citep{qwen2.5-VL} as the base model. The VERA reward uses four components with weights $\lambda_1{=}0.05$, $\lambda_2{=}0.6$, $\lambda_3{=}0.2$, and $\lambda_4{=}0.15$. Full prompt templates are provided in Appendix~\ref{prompts}.

\begin{table}[t]
\setlength{\tabcolsep}{3pt}
\centering
\footnotesize
\begin{tabular}{lccc}
\toprule
\textbf{Model} & \textbf{w/o CoT} & \textbf{w/ CoT} & \textbf{w/ Ours} \\
\midrule
InternVL2.5-2B & 45.7 & 42.7 & \textbf{50.3} {\scriptsize(\textcolor{green!50!black}{+10.1\%})} \\
Qwen2.5-VL-3B  & 55.8 & 48.7 & \textbf{64.3} {\scriptsize(\textcolor{green!50!black}{+15.2\%})} \\
Gemma-3-4B     & 35.2 & 46.2 & \textbf{51.3} {\scriptsize(\textcolor{green!50!black}{+11.0\%})} \\
Qwen2.5-VL-7B  & 58.3 & 63.3 & \textbf{70.9} {\scriptsize(\textcolor{green!50!black}{+12.0\%})} \\
\bottomrule
\end{tabular}
\caption{Backbone-agnostic evaluation.}
\label{plug}
\end{table}

\subsection{Main Results}

\textbf{Performance across Models and Prompting Regimes.}
We evaluate our method on three comic benchmarks under both w/o CoT (direct answering)
and w/ CoT (reasoning-first) settings.
As shown in Table~\ref{dabiao}, our approach consistently outperforms all $\leq$4B models
across benchmarks under both prompting regimes.
Notably, the 3B model equipped with our framework matches or exceeds several 7B MLLMs, demonstrating that enforcing faithful reasoning
does not compromise task performance even under tight capacity constraints.

\noindent\textbf{Faithful Reasoning Evaluation.}
We further evaluate reasoning quality across five benchmarks, comparing our method with three categories of baselines: (1) standard CoT prompting and representative variants, including CoT-SC~\citep{wangself}, ToT~\citep{yao2023tree}, and GoT~\citep{besta2024graph}; (2) representative multimodal CoT methods, including DDCoT~\citep{zheng2023ddcot}, CCoT~\citep{mitra2024compositional}, and Reasoner~\citep{zhang2025improve}; and (3) the recent comic understanding framework LAD~\citep{zhang2025let}. For fair comparison, all methods are implemented on the same backbone, Qwen2.5-VL-3B.

Table~\ref{tab:main_comparison} summarizes the results. Many existing reasoning methods exhibit a trade-off between procedural consistency (CCS) and evidence grounding (USR): they make the rationale align better with the final answer, but often at the cost of introducing more unsupported statements. Our method largely breaks this trade-off, achieving the best overall balance between consistency and grounding across benchmarks. These trends suggest that by improving answer--rationale consistency and reducing unsupported statements, our framework better enhances both reasoning faithfulness and final-task performance.

\noindent\textbf{Plug-in Analysis.}
We evaluate the plug-in capability of our framework by applying it to four representative backbones ranging from 2B to 7B parameters and testing on DeepEval. As shown in Table~\ref{plug}, our method consistently outperforms the stronger of the w/o CoT and w/ CoT baselines across all backbones, with larger gains on smaller models. These results show that our framework transfers well across model scales and serves as an effective plug-in for diverse MLLMs.

\begin{figure*}[t]
  \centering
  \includegraphics[width=1\linewidth]{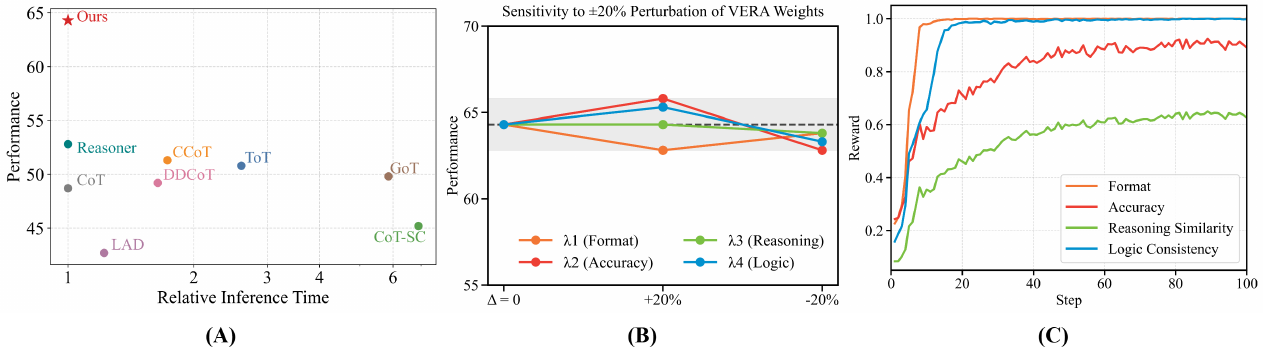}
  \caption{(A) Performance--efficiency trade-off across different reasoning methods. Our method achieves the best overall trade-off, attaining the strongest performance while maintaining the lowest relative inference cost. (B) Sensitivity of VERA to $\pm$20\% perturbations of individual reward weights. The limited performance variation indicates that the proposed reward design is reasonably robust to moderate weight changes. (C) Early-stage dynamics of the four VERA reward components during GRPO training.}
  \label{time_sense}
\end{figure*}

\subsection{Further Analysis}
\label{fur}

\begin{table}[t]
\centering
\footnotesize
\setlength{\tabcolsep}{3pt}
\begin{tabular}{lccc}
\toprule
\textbf{CoT Supervision for GRPO} & \textbf{ACC}$\uparrow$ & \textbf{CCS}$\uparrow$ & \textbf{USR}$\downarrow$ \\
\midrule
gpt-4o-mini-generated CoT & 56.8 & 76.8 & 17.5 \\
\rowcolor{gray!20}
\textbf{MoCoT} & \textbf{60.3} & \textbf{80.6} & \textbf{15.2} \\
\bottomrule
\end{tabular}
\caption{Ablation on supervision source for naive GRPO.}
\label{cot_source_abl}
\end{table}

\noindent\textbf{Failure Patterns and Inference Efficiency.}
To better understand how our framework improves faithful comic reasoning, we compare it with the base model (Qwen2.5-VL-3B) on three representative failure patterns: symbolic misalignment, salient visual cue omission, and satirical target confusion. Representative examples are shown in Figure~\ref{quan}, with additional qualitative cases provided in Appendix~\ref{comqual}. These failures all reflect breakdowns in interpreting and integrating the multiple cues required for comic understanding. In contrast, our method produces reasoning that is more faithful to the visual evidence and better aligned with the symbolic and narrative meaning of the comic. 

Beyond individual examples, we further quantify the frequency of each failure type among incorrect predictions. Figure~\ref{grpo}(B) shows the distribution of failure patterns. Our method consistently reduces all three categories, suggesting that its gains come from mitigating core forms of unfaithful reasoning.

As shown in Figure~\ref{time_sense}(A), our method achieves a favorable performance--efficiency trade-off, attaining the best overall results while incurring the lowest relative inference time among the compared methods. This supports our core design choice of distilling MoCoT into final rationales for GRPO, instead of requiring the model to reproduce the full modular reasoning process at inference time.

\noindent\textbf{Analysis of Faithful Supervision Construction.}
Table~\ref{cot_source_abl} examines whether \textit{stronger-model-generated naive CoT, followed by GRPO, is already sufficient}. The answer is no. Even when generated by gpt-4o-mini, which is already used in MoCoT for the non-Step-2 stages, naive CoT remains less effective than MoCoT supervision under the same GRPO optimization. This suggests that the gain comes not merely from teacher strength, but from more faithful supervision construction.

\noindent\textbf{Analysis of Faithful Reasoning Learning.}

\noindent(1) \textit{Learning Strategy Ablation.} 
Table~\ref{learn_abl} shows a clear trend: under the same MoCoT supervision, performance consistently improves from SFT to naive GRPO, and further to VERA-guided GRPO. This result suggests that faithful supervision alone is not sufficient; how the model learns from such supervision also matters. While MoCoT provides more faithful supervision, VERA further helps the model translate it into more reliable reasoning behavior under self-generated trajectories.

\begin{table}[t]
\centering
\footnotesize
\setlength{\tabcolsep}{3pt}
\begin{tabular}{lccc}
\toprule
\textbf{Learning Strategy} & \textbf{ACC}$\uparrow$ & \textbf{CCS}$\uparrow$ & \textbf{USR}$\downarrow$ \\
\midrule
SFT & 55.8 & 75.6 & 17.8 \\
Naive GRPO & 60.3 & 80.6 & 15.2 \\
\rowcolor{gray!20}
\textbf{VERA}-Guided GRPO & \textbf{64.3} & \textbf{85.9} & \textbf{12.9} \\
\bottomrule
\end{tabular}
\caption{Ablation on learning strategy with MoCoT supervision.}
\label{learn_abl}
\end{table}

\noindent(2) \textit{Reward Component Ablation.}
Table~\ref{tab:vera_ablation} shows that removing any reward term degrades performance, confirming that the four components of VERA play complementary roles. The larger impact of removing $\lambda_3$ or $\lambda_4$ on CCS and USR supports our claim that rationale similarity and logic consistency provide the main faithfulness signals in VERA. By contrast, removing $\lambda_1$ has almost no effect on accuracy and only a marginal effect on CCS and USR, suggesting that format correctness mainly serves an auxiliary role by helping maintain structured outputs.

\noindent(3) \textit{Reward Sensitivity.}
As shown in Figure~\ref{time_sense}(B), when each reward weight is individually adjusted by $\pm 20\%$ from its default setting, the resulting performance changes remain relatively limited, with absolute accuracy variations within 1.5 points. This suggests that VERA is reasonably robust to moderate reward-weight perturbations, and does not require highly precise weight tuning to remain effective in practice.

\noindent\textbf{Optimization Dynamics.}
Figure~\ref{time_sense}(C) shows that the four VERA rewards converge at different rates during early training. The format reward and logic consistency reward rise sharply and saturate early, while the accuracy reward improves more gradually and the reasoning similarity reward increases the slowest. The dynamics therefore indicate that VERA first stabilizes structured and self-consistent generation, and then progressively guides the model toward faithful reasoning patterns under self-generated trajectories.

\begin{table}[t]
\centering
\footnotesize
\setlength{\tabcolsep}{3pt}

\begin{tabular}{l c c c}
\toprule
\textbf{Setting} & \textbf{ACC}$\uparrow$ & \textbf{CCS}$\uparrow$ & \textbf{USR}$\downarrow$ \\
\midrule
\rowcolor{gray!20}
Full \textbf{VERA} ($\lambda_1{+}\lambda_2{+}\lambda_3{+}\lambda_4$) & \textbf{64.3} & \textbf{85.9} & \textbf{12.9} \\

w/o $\lambda_1$ (format correctness)   & \textbf{64.3} & 85.5 & 13.1 \\
w/o $\lambda_2$ (answer accuracy)      & 59.3 & 81.2 & 15.0 \\
w/o $\lambda_3$ (rationale similarity) & 61.3 & 78.9 & 16.8 \\
w/o $\lambda_4$ (logic consistency)    & 56.3 & 73.6 & 16.1 \\

\bottomrule
\end{tabular}
\caption{Ablation on VERA reward component.}
\label{tab:vera_ablation}
\end{table}

\section{Conclusion}

Comic understanding is challenging for small MLLMs because its meaning depends on integrating multiple cues. Naive CoT often collapses this process into a single linear reasoning path too early, causing models to miss critical cues, drift toward unsupported assumptions, and produce explanations that are misaligned with how the answer is derived. We address this problem by formulating comic reasoning as a problem of faithful reasoning, namely reasoning that stays grounded in the relevant visual and textual evidence, integrates it with background knowledge in a symbolically and narratively coherent way, and leads to a final answer supported by the reasoning process itself rather than by post-hoc justification. This requires solving two coupled challenges: constructing supervision that preserves the structure of comic understanding, and learning to follow such supervision reliably when the model must reason from its own generated intermediate steps. Our framework addresses these challenges with MoCoT and VERA.

Across five benchmarks, our framework consistently improves both task performance and reasoning faithfulness while preserving efficiency. More broadly, our findings suggest that this framework may be useful for multimodal reasoning tasks beyond comic understanding, especially those requiring the joint interpretation of multiple cues under non-literal, context-dependent meaning.

\clearpage
\newpage

\bibliographystyle{ACM-Reference-Format}
\bibliography{sec/ref}

\clearpage
\newpage
\appendix

\clearpage
\newpage
\appendix

\section{Algorithm Description}
\label{app:pseudocode}

We provide pseudocode for the main components of our framework and the key evaluation metrics:
(i) MoCoT for faithful supervision construction (Algorithm~\ref{alg:mocot});
(ii) VERA-guided GRPO fine-tuning for faithful reasoning learning (Algorithm~\ref{alg:crft-appendix});
and (iii) the two faithfulness metrics used in our experiments, namely
Counterfactual Consistency Score (CCS; Algorithm~\ref{alg:cas}) and
Unfaithful Statement Rate (USR; Algorithm~\ref{alg:usr}).

\paragraph{CCS and USR (faithfulness metrics).}
Both CCS and USR are computed at the \emph{sample level} (one question-image instance) and
are aggregated by averaging over the evaluation set.
CCS is a binary indicator of \emph{procedural consistency}:
a rationale either exclusively entails the predicted option (while not entailing a counterfactual option) or it does not,
yielding $\mathrm{CCS}\in\{0,1\}$.
USR is a continuous measure of \emph{evidence grounding}:
it computes the fraction of unsupported atomic claims within a rationale,
yielding $\mathrm{USR}\in[0,1]$ (lower is better).

Both metrics are evaluated using the same multimodal judge model,
InternVL3-8B,
to ensure consistency across benchmarks and settings.
The exact judge prompts used for CCS verification
and USR claim extraction are provided in Appendix~\ref{prompts}.
Importantly, these metrics are independent of answer correctness
and focus on the internal alignment between intermediate reasoning
and the final prediction.


\begin{algorithm}
\caption{MoCoT}
\label{alg:mocot}
\begin{algorithmic}[1]
\Require{Comic image $I$, question $Q$}
\Ensure{Final answer $A_o$ with validated rationale}
\State Initialize modules: planner $\mathcal{P}$, executors $\{\mathcal{E}_k\}$, meta-reasoner, and checker $\mathcal{V}$
\Repeat
    \State $\mathcal{Q}_{\text{sub}} \gets \mathcal{P}(I,Q)$ \Comment{Decompose into $K$ typed sub-questions}
    \State $\mathcal{Q}_{\text{sub}} = \{(q_k,t_k)\}_{k=1}^K,\ t_k \in \{\textsc{Visual},\textsc{Symbolic},\textsc{Narrative}\}$
    \State Restrict reasoning space: $\mathcal{Z}_{t_k} \subseteq \mathcal{Z}$ for each type $t_k$
    
    \For{$k=1$ to $K$}
        \State $(r_k, a_k) \gets \mathcal{E}_k(I, q_k; t_k)$ \Comment{Executor produces rationale $r_k$ and provisional answer $a_k$}
    \EndFor
    \State $\mathcal{C}_{\text{sub}} \gets \{(r_k, a_k, t_k)\}_{k=1}^K$ \Comment{Pool of typed sub-results}
    
    \State $\mathrm{DTR} \gets \textsf{Diagnose}(\mathcal{C}_{\text{sub}}, I, Q)$ \Comment{Aggregate evidence into diagnostic rationale}
    \State $(\mathrm{FIR}, A_o) \gets \textsf{Infer}(I,Q;\mathrm{DTR})$ \Comment{Generate final inference rationale and answer}
    \State $A_o' \gets \mathcal{V}(\mathrm{FIR})$ \Comment{Checker validates entailment of the final rationale}
\Until{$A_o' = A_o$}
\State \Return $A_o$
\end{algorithmic}
\end{algorithm}

\begin{algorithm*}
\caption{GRPO Fine-tuning with VERA Reward}
\label{alg:crft-appendix}
\begin{algorithmic}[1]
\Require{Initial policy $\pi_{\omega}^0$, dataset $\mathcal{D}$, reward functions $\{R_{\it format},R_{\it acc},R_{\it rsn},R_{\it logic}\}$ with weights $\{\lambda_i\}$, hyperparameters: $N$ (outer iterations), $M$ (steps per iteration), $\mu$ (GRPO updates), $\epsilon$ (clipping), $\beta$ (KL coefficient)}
\Ensure{Fine-tuned policy $\pi_{\omega}$}
\State $\pi_{\omega} \gets \pi_{\omega}^0$
\For{$n=1,\dots,N$}
    \State $\pi_{\mathrm{ref}} \gets \pi_{\omega}$
    \For{$m=1,\dots,M$}
        \State Sample minibatch $\mathcal{B} \subset \mathcal{D}$
        \State $\pi_{\omega}^{old} \gets \pi_{\omega}$ \Comment{Update old policy}
        \For{each $q \in \mathcal{B}$}
            \State Generate $G$ outputs $\{o_i\}_{i=1}^G \sim \pi_{\omega}^{old}(\cdot|q)$
            \For{$i=1,\dots,G$}
                \State Compute VERA reward:
                \[
                R(o_i) = \lambda_1 R_{\it f}(o_i)+\lambda_2 R_{\it a}(o_i)+\lambda_3 R_{\it r}(o_i)+\lambda_4 R_{\it l}(o_i)
                \]
            \EndFor
            \State Normalize rewards: $\tilde{R}(o_i) = (R(o_i)-\mathrm{mean}(R))/\mathrm{std}(R)$
            \State Set advantages: $\hat{A}_{i,t} \gets \tilde{R}(o_i),\ \forall t \in o_i$
        \EndFor
        \For{$u=1,\dots,\mu$}
            \State Update $\pi_{\omega}$ with gradient coefficient:
            \[
            GC(q,o,t) = \hat{A}_{i,t}+\beta\left(\tfrac{\pi_{\mathrm{ref}}(o_{i,t}\mid o_{i,<t})}{\pi_{\omega}(o_{i,t}\mid o_{i,<t})}-1\right)
            \]
        \EndFor
    \EndFor
\EndFor
\State \Return $\pi_{\omega}$
\end{algorithmic}
\end{algorithm*}

\begin{algorithm}
\caption{Counterfactual Consistency Score (CCS)}
\label{alg:cas}
\begin{algorithmic}[1]
\Require{Image $I$, question $Q$, predicted answer $A_{\text{pred}}$, rationale $R$}
\Ensure{Procedural consistency indicator $\mathrm{CCS}\in\{0,1\}$}

\State $A_{\text{cf}} \gets \textsf{SelectCounterfactual}(A_{\text{pred}}, Q)$
\Comment{Choose a plausible alternative answer}

\State $e^{+} \gets \textsf{Entail}(I,Q,R,A_{\text{pred}})$
\State $e^{-} \gets \textsf{Entail}(I,Q,R,A_{\text{cf}})$

\State $\mathrm{CCS} \gets \mathbb{I}[\,e^{+}=1 \wedge e^{-}=0\,]$
\State \Return $\mathrm{CCS}$
\end{algorithmic}
\end{algorithm}

\begin{algorithm}
\caption{Unfaithful Statement Rate (USR)}
\label{alg:usr}
\begin{algorithmic}[1]
\Require{Image $I$, question $Q$, rationale $R$}
\Ensure{Evidence-grounding score $\mathrm{USR}\in[0,1]$}

\State $\mathcal{C} \gets \textsf{ExtractClaims}(R)$
\Comment{Decompose rationale into atomic statements}

\State $u \gets 0$
\For{each claim $c \in \mathcal{C}$}
    \If{\textsf{Grounded}(c, I, Q)=0}
        \State $u \gets u + 1$
    \EndIf
\EndFor

\State $\mathrm{USR} \gets u / |\mathcal{C}|$
\State \Return $\mathrm{USR}$
\end{algorithmic}
\end{algorithm}

\section{Related Work}\label{app:related}

\paragraph{Comic Understanding as Cross-Cue Multimodal Reasoning.}
Comics are a distinctive multimodal narrative form whose meaning often emerges from the joint interpretation of visual content, textual expressions, and broader social or cultural knowledge, rather than from a single literal reading. This makes comic understanding fundamentally different from standard visual question answering, as successful interpretation often requires resolving incongruity, implicit intent, symbolism, and narrative context through cross-cue integration. Recent benchmarks have further established comic and meme understanding as a challenging testbed for MLLMs, covering humor understanding, implicature, satire, and culturally grounded interpretation \citep{hessel2023androids,hwang2023memecap,yang2024can,liu2024ii,hu2024cracking,zhang-etal-2025-mllms}. While these works have clarified the difficulty of the task and revealed substantial gaps between MLLMs and human performance, they have primarily focused on evaluation and capability analysis, leaving open how the reasoning process itself can be better supported, especially for small MLLMs.

\paragraph{CoT Reasoning in MLLMs and Its Structural Limits.}
Chain-of-Thought (CoT) prompting has become a standard approach for eliciting intermediate reasoning in LLMs and MLLMs \citep{wei2022chain,kojima2022large}. Prior work has extended CoT to multimodal settings through structured prompting or intermediate representations, such as modular decomposition \citep{zheng2023ddcot} and scene-graph-based reasoning \citep{mitra2024compositional}. More broadly, reasoning research has explored richer inference-time strategies such as self-consistency, tree-structured search, graph-structured reasoning, critique, and metacognitive planning \citep{wangself,yao2023tree,besta2024graph,du2023improving,radhakrishnan2023question,bai2025mp}. These studies collectively suggest that reasoning quality depends not only on whether intermediate steps are elicited, but also on how they are organized and constrained. However, most existing approaches still assume that making reasoning explicit is generally helpful, without asking whether a single linear reasoning chain is actually appropriate for tasks like comic understanding, where multiple cues and candidate interpretations often need to remain active until they can be jointly resolved. This limitation becomes especially important in small MLLMs, which are more vulnerable to reasoning drift and premature commitment under naive CoT prompting.

\paragraph{Faithful Supervision Construction for Reasoning.}
A related line of work seeks to improve reasoning by improving supervision itself, including rationale construction, explicit decomposition, process supervision, and verifier-guided filtering \citep{lightman2023let,radhakrishnan2023question}. These works show that the structure of supervision can strongly affect what reasoning patterns a model ultimately learns. Recent multimodal reasoning efforts have similarly explored verifiable or multi-perspective supervision, for example through large-scale CoT datasets and verifier designs that jointly assess logical consistency and visual grounding \citep{luo2025ursa}. However, such approaches are typically developed for general reasoning settings and do not explicitly target the interpretive structure of comics, where implicit assumptions, symbolic cues, and contextual knowledge are tightly entangled. In contrast, our work focuses on constructing supervision that is itself more faithful to the multi-cue and context-sensitive nature of comic reasoning.

\paragraph{Reward-Based Reasoning Learning in MLLMs.}
Beyond supervised learning, recent work has explored reinforcement learning for multimodal reasoning, motivated by the advantage of optimizing models directly on their own generated reasoning trajectories rather than only on ground-truth histories \citep{keneshloo2019deep,shao2024deepseekmath,guo2025deepseek}. In MLLMs, representative efforts include reinforcement-based reasoning with modality bridging, verifier-guided optimization, and diversity-aware multi-path reasoning \citep{huang2025vision,shi2025multimodal,luo2025ursa}. These methods have advanced multimodal reasoning optimization, but they often rely on large-scale training, strong base models, or task-specific supervision. More importantly, optimizing only for outcome correctness does not by itself ensure that the reasoning process remains faithful and well grounded. For small MLLMs, there is also a practical requirement to improve reasoning quality without reproducing heavy procedural reasoning at inference time. Our work is most closely related to this line, but differs in explicitly framing the problem as one of faithful comic reasoning and addressing it through two coupled components: faithful supervision construction and faithfulness-aware reward-based learning.

\section{Prompt List}\label{prompts}

We provide the exact system prompts used in our experiments. Specifically, Table~\ref{app:s1}, ~\ref{app:s2}, ~\ref{app:s3.1}, and ~\ref{app:s3.2} correspond to the prompts for Step~1 (Subgoal Planning), Step~2 (Localized Execution), and Step~3 (Meta-Reasoning and Verification) in the MoCoT pipeline. In addition, Table~\ref{app:p_grpo} presents the system prompt used for VERA-guided GRPO fine-tuning, which enforces structured output formatting. Finally, Tables~\ref{app:wocot} and \ref{app:wcot} provide the prompts employed in evaluating MLLMs without and with chain-of-thought reasoning, respectively.

For multiple-choice reasoning tasks, all methods use a standardized option-selection prompt that requires explicit answer identification (Table~\ref{app:p_grpo}).
For open-ended settings (e.g., MemeCap), we adopt a constrained reasoning–explanation prompt that enforces visual grounding and concise intent description (Table~\ref{app:memecap}).

Finally, the two faithfulness metrics introduced in this work—Counterfactual Consistency Score (CCS) and Unfaithful Statement Rate (USR)—are implemented using fixed multimodal judge prompts.
The prompt used for CCS entailment verification is provided in Table~\ref{app:prompt_CCS}, while the prompt used for USR claim-level grounding evaluation is shown in Table~\ref{app:prompt_usr}.

\begin{table}
\centering
\caption{Prompt used in Step~1 (Subgoal Planning) of the MoCoT pipeline.}
\begin{tabular}{p{0.92\linewidth}}
\toprule
You will be shown an image and a related question. Do not attempt to answer the question. Instead, analyze the question in the context of the image, focusing on what makes it semantically complex, ambiguous, or rich in interpretation. Then, decompose it into a small number of \textbf{independent sub-questions} (up to 4), each targeting a different aspect that would help a model better understand and reason about the original question.
These sub-questions should:
\begin{itemize}
    \item Reflect the implicit knowledge, reasoning steps, or background assumptions required to fully understand the question and its connection to the image.
    \item Avoid directly referencing answer options.
    \item Be useful for guiding deeper interpretation of the comic’s meaning, narrative structure, symbolism, or emotional content.
\end{itemize}
Format your output as:

\begin{lstlisting}[language=json]
{
  "cot": "<your step-by-step reasoning about the question and how the image informs it>",
  "sub_questions": [
    "<sub-question 1>",
    "<sub-question 2>",
    "... (up to 4)"
  ]
}
\end{lstlisting}

Only output the JSON. You may refer to the image to inform your reasoning, but do not describe the image or answer the question. \\
\bottomrule
\end{tabular}
\label{app:s1}
\end{table}

\begin{table}
\centering
\caption{Prompt used in Step~2 (Localized Execution) of the MoCoT pipeline.}
\begin{tabular}{p{0.92\linewidth}}
\toprule
You will be shown an image and a related question. Your task is to answer the question using a Chain of Thought (CoT) approach, grounded in the image.
\begin{itemize}
    \item First, examine the image and identify relevant visual elements (e.g., objects, characters, expressions, actions, spatial relationships).
    \item Then, reason through the question step by step, referencing the image as needed.
    \item Finally, answer the question directly.
\end{itemize}
Your output must be in the following JSON format:

\begin{lstlisting}[language=json]
{
  "cot": "<your step-by-step reasoning based on the image and question>",
  "answer": "<final answer here>"
}
\end{lstlisting}

Now, analyze the image and answer the question with a clear CoT reasoning process in the specified JSON format. \\
\bottomrule
\end{tabular}
\label{app:s2}
\end{table}

\begin{table*}
\centering
\caption{Prompt used in Step~3 (Meta-Reasoning) of the MoCoT pipeline.}
\begin{tabular}{p{1\linewidth}}
\toprule
\textbf{Two-Stage Visual Reasoning: Interpret the Deep Meaning of a Cartoon}

You are provided with:
\begin{itemize}
  \item A \textbf{cartoon image} (`image\_path')
  \item A \textbf{multiple-choice question} asking which of the provided options (e.g., A, B, C, …) best expresses the cartoon’s deep meaning
  \item A set of \textbf{sub-questions and sub-answers} exploring visual, symbolic, or thematic aspects of the image
\end{itemize}

Your task involves two distinct reasoning stages:

\textbf{Stage 1 — cot1: Critically Evaluate Sub-Answers}  
Do not try to answer the main question yet. For each sub-question and its answer:
\begin{itemize}
  \item Assess whether the answer is accurate, coherent, visually grounded, and symbolically insightful.
  \item Point out strong insights (e.g., symbolism, emotional interpretation).
  \item Point out weak points (e.g., vagueness, factual errors, irrelevance).
\end{itemize}
Summarize in a concise paragraph or bullet list per sub-answer. The goal is to diagnose the quality of intermediate reasoning, not to solve the problem.

\textbf{Stage 2 — cot2: Independent Deep Reasoning and Final Choice (Informed by cot1)}  
Now interpret the cartoon from the image itself, making an independent judgment. Steps:
\begin{enumerate}
  \item Describe the image explicitly (main objects, actions, tone, key symbols).
  \item Interpret the symbolism and theme (message, human values, societal critique).
  \item Compare all answer choices: explain matches and mismatches.
\end{enumerate}
Finally, give your best answer.

\textbf{Final Output Format:}
\begin{lstlisting}[language=json]
{
  "cot1": "Your structured evaluation of the sub-answers.",
  "cot2": "Your independent reasoning and answer justification.",
  "answer": "Your final choice (e.g., A, B, C, D, or other label)"
}
\end{lstlisting}

\textbf{Example Output:}
\begin{lstlisting}[language=json]
{
  "cot1": "1. The answer to sub-question 1 accurately identifies the image's central element - a businessman climbing over others. It is visually grounded and symbolically points to social hierarchy.
      2. The answer to sub-question 2 misses the emotional tone -- the despair of those stepped on. It's a surface-level description without symbolic insight. 
      3. Sub-answer 3 insightfully connects the broken ladder to systemic inequality -- a strong symbolic interpretation.",
  "cot2": "The image depicts a businessman climbing a ladder made of people. Those below appear crushed, while he ascends smugly. The exaggerated expressions emphasize exploitation. Symbolically, the cartoon critiques how success in capitalism often rests on the suffering of others.
      A: Suggests hard work pays off -- doesn't fit the exploitative theme.
      B: Argues society rewards the clever -- also fails to address the cruelty shown.
      C: Says 'one's success is built on others' pain' -- this directly reflects the image's symbolism.
      D: Suggests individualism is key -- irrelevant to the collective suffering shown. C is the best fit.",
  "answer": "C"
}
\end{lstlisting}

 \\
\bottomrule
\end{tabular}
\label{app:s3.1}
\end{table*}

\begin{table*}
\centering
\caption{Prompt used in Step~3 (Verification) of the MoCoT pipeline.}
\begin{tabular}{p{0.92\linewidth}}
\toprule
You are a logical critique model tasked with post-hoc evaluation and revision of a reasoning paragraph (`cot2') that aims to justify the selection of one of several options (e.g., A, B, C, D) in response to a visual question. \textbf{You will NOT see the image}, only the textual reasoning.

\textbf{Objectives:}
\begin{enumerate}
  \item Determine if the original `cot2' logically supports the given final answer.
  \item If it does not, return a corrected version of `cot2'.
\end{enumerate}

\textbf{Output Format:}  
Respond with a valid JSON object, enclosed in a markdown code block, like this:
\begin{lstlisting}[language=json]
{
  "Matched Answer": "A",
  "Is Consistent": true,
  "Justification": "The reasoning supports the final answer.",
  "Corrected CoT2": "The revised reasoning here."
}
\end{lstlisting}

Do not include anything outside the code block. \\
\bottomrule
\end{tabular}
\label{app:s3.2}
\end{table*}

\begin{table*}
\centering
\caption{Prompt used for GRPO reinforcement learning fine-tuning,
which also serves as a reasoning-inductive prompt for all reasoning-based methods.}

\begin{tabular}{p{0.92\linewidth}}
\toprule
A conversation between User and Assistant. The user asks a multiple-choice question, and the Assistant solves it. The assistant first thinks about the reasoning process in the mind and then provides the user with the answer. The reasoning process and answer are enclosed within \texttt{<REASONING></REASONING>} and \texttt{<ANSWER></ANSWER>} tags, respectively, i.e., \texttt{<REASONING> reasoning process here </REASONING><ANSWER> answer option label here </ANSWER>} 
\\
\bottomrule
\end{tabular}
\label{app:p_grpo}
\end{table*}

\begin{table*}
\centering
\caption{Prompt used for open-ended meme and comic understanding on MemeCap.}
\label{app:memecap}

\begin{tabular}{p{0.92\linewidth}}
\toprule
A conversation between User and Assistant. The user asks an open-ended question about a meme or comic.
You MUST output in exactly this format: \\
\texttt{<REASONING>...</REASONING><ANSWER>...</ANSWER>} \\[4pt]

Inside \texttt{<REASONING>}, use EXACTLY the following template with TWO lines (no additional steps): \\
\texttt{CUES:} 2--4 short phrases describing key visible elements or clearly readable text. \\
\texttt{INTENT:} one sentence explaining the joke or intent based ONLY on CUES. \\[4pt]

Rules: \\
1) \texttt{<ANSWER>} must be ONE concise sentence explaining the joke. \\
2) \texttt{<ANSWER>} must reuse at least one phrase from CUES. \\
3) Do NOT claim ``the text says $\ldots$'' unless the exact text is clearly readable. \\
4) Forbidden words in BOTH \texttt{<REASONING>} and \texttt{<ANSWER>}: societal, norms, expectations, resilience, redemption. \\
5) Before outputting, self-check and rewrite to remove any forbidden words. \\
\bottomrule
\end{tabular}
\end{table*}

\begin{table*}
\centering
\caption{Prompt used for MLLM evaluation without CoT.}
\begin{tabular}{p{0.92\linewidth}}
\toprule
You are a helpful assistant for image-based reasoning.  
You will be given an image and a related multiple-choice question. Your task is to examine the image, understand the question and options, and then identify the correct answer.  

Respond in \textbf{strict JSON format}, with only one field:  
- "answer": a list that contains only your final answer as a string — specifically, the correct option label (e.g., "A", "B", "C", or "D"). It should NOT include explanation or restate the option text.  

\textbf{Output exactly one JSON object, and nothing else — no comments, no code blocks, no additional text.}  

Here is the required format:  

\begin{lstlisting}[language=json]
{
  "answer": ["<option label here>"]
}
\end{lstlisting}

If the answer is not certain, make your best inference and still produce a valid JSON object with your final answer.
\\
\bottomrule
\end{tabular}
\label{app:wocot}
\end{table*}

\begin{table*}
\centering
\caption{Prompt used for MLLM evaluation with CoT.}
\begin{tabular}{p{0.92\linewidth}}
\toprule
You will be given an image and a related multiple-choice question. Your task is to examine the image, understand the question and options, and then reason step by step before arriving at the final answer.

Respond in strict JSON format, with two fields:
- "cot": a detailed step-by-step explanation showing your reasoning based on visual elements and the question and options.
- "answer": a list that contains only your final answer (e.g., ["A"]).

\textbf{Output exactly one JSON object, and nothing else.}

Your output must be in the following JSON format:
\begin{lstlisting}[language=json]
{
  "cot": "<your step-by-step reasoning based on the image and question>",
  "answer": "<final answer here>"
}
\end{lstlisting}

\\
\bottomrule
\end{tabular}
\label{app:wcot}
\end{table*}

\begin{table*}
\centering
\caption{Prompt used for Counterfactual Consistency Score (CCS) evaluation.}
\begin{tabular}{p{0.92\linewidth}}
\toprule
You are a strict multimodal natural language inference (NLI) judge for evaluating reasoning faithfulness.

You are given:
\begin{itemize}
  \item an image,
  \item a multiple-choice question with options,
  \item a claimed answer (one option letter),
  \item and a rationale produced by a model.
\end{itemize}

Your task is to determine whether the rationale \emph{logically entails} the claimed answer,
based \textbf{only} on what is visible in the image and the provided question and options.

\textbf{Evaluation rules:}
\begin{itemize}
  \item Do \emph{not} reward fluent language or plausible storytelling.
  \item Penalize unsupported assumptions, speculation, or missing visual evidence.
  \item If the rationale could also support a different option, then it does \emph{not} entail the claimed answer.
  \item Judge entailment, not answer correctness.
\end{itemize}

\textbf{Output format (JSON only):}
\begin{lstlisting}[language=json]
{
  "entails": 0 or 1,
  "confidence": 0.0 to 1.0,
  "support_spans": ["<=20 words quoted from the rationale", "..."],
  "counter_spans": ["<=20 words indicating ambiguity or unsupported leaps", "..."]
}
\end{lstlisting}

Output exactly one JSON object and nothing else.
\\
\bottomrule
\end{tabular}
\label{app:prompt_CCS}
\end{table*}

\begin{table*}
\centering
\caption{Prompt used for Unfaithful Statement Rate (USR) evaluation.}
\begin{tabular}{p{0.92\linewidth}}
\toprule
You are a strict multimodal faithfulness judge.

Your goal is to compute the \emph{Unfaithful Statement Rate (USR)}
for a given rationale.

You are given:
\begin{itemize}
  \item an image,
  \item a question with options,
  \item and a rationale produced by a model.
\end{itemize}

\textbf{Task:}
\begin{itemize}
  \item Extract \textbf{exactly 5} atomic claims from the rationale
        (each claim must be \(\leq\) 10 words).
  \item If fewer than 5 claims are present, pad with
        \texttt{\{"id":k,"t":"<NONE>","u":0\}}.
  \item For each claim, set \texttt{u = 1} if it is an unsupported assumption,
        speculation, or inferred intent (e.g., motives, societal meaning)
        that is \textbf{not grounded} in the image or the question/options.
  \item Otherwise, set \texttt{u = 0}.
\end{itemize}

USR is defined as:
\[
\mathrm{USR} = \frac{\#\{u = 1\}}{N},
\]
where \(N = 5\).

\textbf{Output format (JSON only):}
\begin{lstlisting}[language=json]
{
  "claims": [
    {"id": 1, "t": "...", "u": 0},
    {"id": 2, "t": "...", "u": 1}
  ],
  "USR": 0.0
}
\end{lstlisting}

Output exactly one JSON object and nothing else.
\\
\bottomrule
\end{tabular}
\label{app:prompt_usr}
\end{table*}

\section{Additional Experimental Results}

\subsection{Full Results for Figure~\ref{moti}(A)}
\label{app:cii-bench-full}

For completeness, we report the full numerical results corresponding to 
Figure~\ref{moti}(A), which illustrates the effect of naive CoT prompting 
on \textbf{CII-Bench}. While the main paper shows the accuracy change in 
aggregate, Tables~\ref{tab:small-mlmms} and \ref{tab:large-mlmms} provide 
the detailed results for Small and Large MLLMs, respectively. 
As can be seen, naive CoT prompting often leads to performance drops, 
especially for smaller models.

\begin{table*}
\centering
\begin{tabular}{l c c c}
\toprule
\textbf{Model} & \textbf{\#Params} & \textbf{w/ CoT} & \textbf{w/o CoT ($\Delta$)} \\
\midrule
Mono-InternVL         & 2B  & 10.7 & 22.5 (+11.8) \\
Ovis2                 & 2B  & 26.8 & 36.3 (+9.5) \\
InternVL2.5           & 2B  & 33.3 & 33.6 (+0.3) \\
Qwen2.5-VL            & 3B  & 36.2 & 41.8 (+5.6) \\
Phi-3.5               & 4B  & 22.1 & 33.1 (+11.0) \\
Qwen2-VL              & 7B  & 50.0 & 49.6 (-0.4) \\
LLaVA1.6              & 7B  & 29.0 & 30.2 (+1.2) \\
InternLM-XComposer-2.5& 7B  & 32.6 & 32.6 (+0.0) \\
Qwen2.5-VL            & 7B  & 45.8 & 48.1 (+2.3) \\
Idefics2$^{*}$              & 8B  & 33.3 & 36.3 (+3.0) \\
MiniCPM-V2.5$^{*}$          & 8B  & 35.8 & 40.4 (+4.6) \\
MiniCPM-V2.6$^{*}$          & 8B  & 38.9 & 45.0 (+6.1) \\
InternVL2$^{*}$             & 8B  & 47.9 & 53.1 (+5.2) \\
InternVL3             & 8B  & 47.7 & 50.7 (+3.0) \\
Qwen-VL-Chat$^{*}$          & 9B  & 34.0 & 34.3 (+0.3) \\
GLM-4V$^{*}$                & 9B  & 49.1 & 50.3 (+1.2) \\
\bottomrule
\end{tabular}
\caption{Accuracy of \textbf{Small} ($\leq$ 15B) MLLMs with and without CoT prompting on the \textbf{CII-Bench} benchmark. 
$\Delta$ = (w/o CoT – w/ CoT). $^{*}$Results reported from \citep{zhang-etal-2025-mllms}.}
\label{tab:small-mlmms}
\end{table*}

\begin{table*}
\centering
\begin{tabular}{l c c c}
\toprule
\textbf{Model} & \textbf{\#Params} & \textbf{w/ CoT} & \textbf{w/o CoT ($\Delta$)} \\
\midrule
CogVLM2               & 19B & 22.2 & 20.3 (-1.9) \\
CogVLM2-Chinese-Chat$^{*}$       & 19B & 42.6 & 43.4 (+0.8) \\
Gemma3                & 27B & 39.1 & 39.1 (+0.0) \\
Qwen2.5-VL            & 32B & 53.7 & 56.2 (+2.5) \\
LLaVA1.6$^{*}$              & 34B & 44.5 & 46.0 (+1.5) \\ 
InternVL3             & 38B & 52.8 & 52.7 (-0.1) \\
InternVL2$^{*}$             & 40B & 57.6 & 57.9 (+0.3) \\
LLaVA1.6$^{*}$              & 72B & 45.3 & 48.0 (+2.7) \\
Qwen2-VL$^{*}$              & 72B & 62.1 & 64.4 (+2.3) \\
InternVL2$^{*}$             & 76B & 52.6 & 52.9 (+0.3) \\

\bottomrule
\end{tabular}
\caption{Accuracy of \textbf{Large} ($>$ 15B) MLLMs with and without CoT prompting on the \textbf{CII-Bench} benchmark. 
$\Delta$ = (w/o CoT – w/ CoT). $^{*}$Results reported from \citep{zhang-etal-2025-mllms}.}
\label{tab:large-mlmms}
\end{table*}

\subsection{Complete Qualitative Comparisons}\label{comqual}

In Figure~\ref{quan}, we highlighted three representative failure modes of Qwen-2.5VL-3B: (A) satirical target confusion, (B) symbolic misalignment, and (C) salient cue omission. In this section, we provide the complete set of responses corresponding to Figure~1, including both Qwen and ours. Moreover, we further illustrate each failure type with additional examples: satirical target confusion with Figures~\ref{apppic3} and~\ref{apppic4}, symbolic misalignment with Figures~\ref{apppic1} and~\ref{apppic6}, and salient cue omission with Figures~\ref{apppic2} and~\ref{apppic5}.



\begin{table*}
\centering
\caption{Qualitative comparison on cartoon question 1.}
\label{apppic1}
\setlength{\tabcolsep}{4pt} 
\begin{tabular}{@{}m{0.45\linewidth} m{0.45\linewidth}@{}}
\toprule
\multicolumn{1}{c}{\textbf{Image}} & \multicolumn{1}{c}{\textbf{Question}} \\
\midrule
\includegraphics[width=\linewidth]{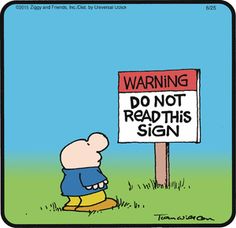} &
\small
A. As illustrated at the top of the frame, this cartoon wants to tell the viewer: be careful with your words. If you express what you think in your head too directly, words will become a sharp weapon to kill you.

B. In some cases, people read the warnings as required, only to find that the warnings themselves were self-denying. This can be confusing. So sometimes the rules themselves contradict each other. We should learn to judge for ourselves.

C. The comic makes fun of the fact that many people are more oblivious towards common knowledge than people expect, and would perform stupid actions on common and easy tasks.

D. The comic strip conveys a deeper meaning related to the divergent outcomes resulting from different choices and behaviors. The caricature also serves as a reminder that our actions have consequences and that making positive choices can lead to personal growth, fulfillment, and a vibrant life, while negative habits can lead to stagnation and a decline in overall well-being.
\\
\toprule
\multicolumn{1}{c}{\textbf{Qwen2.5VL-3B}} & \multicolumn{1}{c}{\textbf{Ours }} \\
\midrule
\small\raggedright
The image shows a cartoon character standing next to a sign that reads "WARNING: DO NOT READ THIS SIGN." This sign is placed in a grassy area with a blue sky in the background. The character appears to be looking at the sign, suggesting that the viewer is meant to pay attention to the message. The sign itself is a play on words, as it is a warning sign, but the text on the sign is actually a warning against reading the sign. This creates a humorous and ironic situation where the warning is not what it seems. The deep meaning of the image can be interpreted as a commentary on the importance of being cautious with our words and actions, as the sign itself is a warning against reading it. The character's reaction to the sign, combined with the irony of the message, suggests that the viewer should be careful with theown words and actions, as they can have unintended consequences.

Final answer: A \textcolor{red}{\ding{55}} 
&
\small\raggedright
The image shows a cartoon character standing next to a sign that reads "WARNING DO NOT READ THIS SIGN." This suggests a paradoxical situation where the warning itself is contradictory. The cartoon humorously highlights the tension between the warning and the instruction to not read it, implying a deeper commentary on the complexity of warnings and the human tendency to ignore or misinterpret them. The cartoon conveys a critique of how people often fail to heed clear warnings, reflecting on the self-denying nature of warnings. Option A focuses on the danger of direct communication, but it doesn't capture the cartoon's critique of warnings. Option B directly addresses the contradiction in warnings, aligning well with the cartoon's message. Option C emphasizes obliviousness to common knowledge, but it misses the core theme of the cartoon's critique. Option D discusses the consequences of choices, but it doesn't specifically relate to the cartoon's focus on warnings. Therefore, option B is the most suitable choice as it encapsulates the cartoon's deeper meaning about the self-denying nature of warnings.

Final answer: B \textcolor{green}{\ding{51}} 

\end{tabular}
\end{table*}


\begin{table*}
\centering
\caption{Qualitative comparison on cartoon question 2.}
\label{apppic2}
\setlength{\tabcolsep}{4pt} 
\begin{tabular}{@{}m{0.45\linewidth} m{0.45\linewidth}@{}}
\toprule
\multicolumn{1}{c}{\textbf{Image}} & \multicolumn{1}{c}{\textbf{Question}} \\
\midrule
\includegraphics[width=\linewidth]{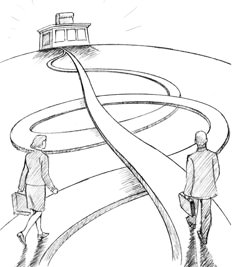} &
\scriptsize
A. Through the echo of the behavior of the mother in 1995 and the behavior of the mother in 2015, the author of the cartoon expressed that with the changes of the times, parents' educational concepts are likely to fall into a strange cycle.

B. The deep meaning conveyed by the picture relates to the unfair treatment of men and women in the workplace. It symbolizes the different challenges and obstacles faced by individuals based on their gender. The direct path represents the easier and smoother career progression often enjoyed by men, while the winding path signifies the complex and difficult journey that women often have to navigate to achieve the same level of success.

C. Our whole life is like being on a conveyor belt. We are always pursuing various goals, yet we never stop chasing them. As we grow older, we start to escape from the threat of death. This cartoon sketches out the essence of life and the dilemma of human beings through a simple scene. We are always running towards fame, fortune, power, etc., but when we reach the finish line, we may only be left with emptiness and endless anxiety.

D. The picture shows that by choosing the right way, you can reach the endpoint, while if you choose the wrong way, you will only get farther and farther away from the endpoint.
\\
\toprule
\multicolumn{1}{c}{\textbf{Qwen2.5VL-3B}} & \multicolumn{1}{c}{\textbf{Ours }} \\
\midrule
\scriptsize\raggedright
The image shows two people walking on a winding path that leads to a tower. The path is marked with concentric circles, suggesting a journey or a path to a destination. Option A: This option suggests that the image is about the changes in parents' educational concepts over time. However, the image does not provide any information about parents' educational concepts or their evolution over time. It focuses more on the path and the tower, which are not directly related to this concept. Option B: This option suggests that the image is about the unfair treatment of men and women in the workplace. The winding path could symbolize the challenges faced by women, while the direct path could represent the easier career progression for men. However, the image does not provide any specific context about gender equality or workplace issues. Option C: This option suggests that the image is about the essence of life and the dilemma of human beings. The winding path could represent the pursuit of various goals, while the tower could symbolize the end or destination. However, the image does not provide any specific context about life or human dilemmas. Option D: This option suggests that the image is about the importance of choosing the right path. The winding path could represent a wrong path, while the direct path could represent the right path. This option aligns well with the visual elements of the image, where the winding path leads to a tower, suggesting a journey or a path to a destination. Based on this analysis, the most suitable deep meaning for the image is Option D: The picture shows that by choosing the right way, you can reach the endpoint, while if you choose the wrong way, you will only get farther and farther away from the endpoint.

Final answer: D  \textcolor{red}{\ding{55}} 

&
\small\raggedright
The image depicts two individuals walking on different paths, one straight and one winding, towards a destination. This suggests a metaphor for different career paths and challenges faced by men and women. The winding path represents the complexities and obstacles women often encounter, while the straight path symbolizes easier career progression for men. The cartoon conveys the theme of gender inequality in professional environments, highlighting the different challenges faced by individuals based on their gender. Option A focuses on the changes in parental educational concepts over time, which does not align with the image's focus on gender inequality. Option B directly addresses the unfair treatment of men and women in the workplace, aligning well with the image's message. Option C emphasizes the essence of life and human pursuit, but it doesn't capture the specific gender dynamics depicted in the image. Option D suggests a choice between right and wrong paths, but it misses the underlying theme of gender inequality. Therefore, option B is the most suitable choice as it encapsulates the deep meaning of the image.

Final answer: B \textcolor{green}{\ding{51}} 

\end{tabular}
\end{table*}


\begin{table*}
\centering
\caption{Qualitative comparison on cartoon question 3.}
\label{apppic3}
\setlength{\tabcolsep}{4pt} 
\begin{tabular}{@{}m{0.45\linewidth} m{0.45\linewidth}@{}}
\toprule
\multicolumn{1}{c}{\textbf{Image}} & \multicolumn{1}{c}{\textbf{Question}} \\
\midrule
\includegraphics[width=\linewidth]{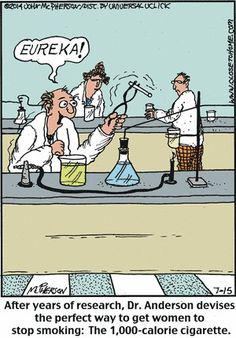} &
\small

A. This cartoon tells us that doing anything requires not only passion but also professional knowledge and careful preparation.

B. Calories are the amount of heat and energy contained in food, which goes against the common pursuit of women in today's times to be slim and build a good body. People tend to choose foods, drinks, etc. with low calorie content in order to lose weight, and even keep their daily calorie intake within a certain limit.

C. This picture shows that many great scientific discoveries often come from experiments with eccentricity.We should not be afraid to question and ridicule to make great discoveries.

D. The comic strip prompts reflection of the status of teachers today, and the shifting landscape of education, parental involvement, and the recognition that a student's self-confidence and personal growth should be valued alongside academic performance.

\\
\toprule
\multicolumn{1}{c}{\textbf{Qwen2.5VL-3B}} & \multicolumn{1}{c}{\textbf{Ours }} \\
\midrule
\scriptsize\raggedright
The image depicts a humorous scenario where a scientist, Dr. Anderson, invents a 1,000-calorie cigarette as a way to get women to stop smoking. This is a satirical take on the idea that scientific discoveries can sometimes be absurd or impractical. Let's analyze each option in relation to the image: A. This cartoon tells us that doing anything requires not only passion but also professional knowledge and careful preparation. - This option is not directly related to the image. The image is more about the absurdity of the invention rather than the preparation required. B. Calories are the amount of heat and energy contained in food, which goes against the common pursuit of women in today's times to be slim and build a good body. People tend to choose foods, drinks, etc. with low calorie content in order to lose weight, and even keep their daily calorie intake within a certain limit. - This option is also not directly related to the image. The image is about a 1,000-calorie cigarette, not about calorie intake or weight loss. C. This picture shows that many great scientific discoveries often come from experiments with eccentricity. We should not be afraid to question and ridicule to make great discoveries. - This option is the most suitable deep meaning for the image. The image humorously depicts a scientist inventing a 1,000-calorie cigarette, which is an absurd and eccentric idea. This aligns with the idea that great scientific discoveries can sometimes come from unconventional and eccentric experiments. D. The comic strip prompts reflection of the status of teachers today, and the shifting landscape of education, parental involvement, and the recognition that a student's self-confidence and personal growth should be valued alongside academic performance. - This option is not related to the image. The image is about a scientist inventing a 1,000-calorie cigarette, not about the status of teachers or education. Therefore, the most suitable deep meaning for the image is option C.

Final answer: C \textcolor{red}{\ding{55}} 
&
\small\raggedright
The cartoon depicts Dr. Anderson inventing a 1,000-calorie cigarette as a way to get women to stop smoking. This suggests a humorous take on the societal pressure to be thin and the pursuit of weight loss. The cartoon implies a critique of societal norms and the pursuit of thinness, highlighting the absurdity of creating a product to counteract these pressures. The deep meaning of the cartoon revolves around the societal expectations and the humorous commentary on the pursuit of thinness. Option A focuses on the need for passion and professional knowledge, which doesn't directly relate to the cartoon's critique of societal norms. Option B emphasizes the common pursuit of slimness, aligning well with the cartoon's critique of societal expectations. Option C discusses the nature of scientific discoveries, but it misses the satirical tone of the cartoon. Option D addresses the status of teachers and education, which is not relevant to the cartoon's message. Therefore, option B is the most suitable choice as it captures the cartoon's critique of societal expectations related to weight and thinness.

Final answer: B \textcolor{green}{\ding{51}} 

\end{tabular}
\end{table*}


\begin{table*}
\centering
\caption{Qualitative comparison on cartoon question 4.}
\label{apppic4}
\setlength{\tabcolsep}{4pt} 
\begin{tabular}{@{}m{0.45\linewidth} m{0.45\linewidth}@{}}
\toprule
\multicolumn{1}{c}{\textbf{Image}} & \multicolumn{1}{c}{\textbf{Question}} \\
\midrule
\includegraphics[width=\linewidth]{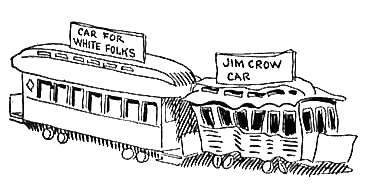} &
\small

A. The cartoon satirizes that the I.R. law is a road to nowhere, yet the government insists on enacting and enforcing it.

B. This cartoon satirizes many people in society who hope for change, but every one of them refuses to change themselves. 

C. Individuals lose their unique identities and become part of a homogeneous group. This portrayal reflects the potential loss of individuality and creativity in standardized educational systems, where diversity and personal expression may be suppressed in favor of uniformity. The comic strip highlights the societal pressure to conform and raises questions about the impact of institutional structures on personal identity and freedom of expression. 

D. It is a sign of racism, because blacks are treated differently, not only are they not allowed to be with whites, but they are kept in worse conditions.

\\
\toprule
\multicolumn{1}{c}{\textbf{Qwen2.5VL-3B}} & \multicolumn{1}{c}{\textbf{Ours }} \\
\midrule
\small\raggedright
The image depicts a train with two separate compartments labeled 'Car for White Folks' and 'Jim Crow Car,' suggesting a division based on race. This satirical representation implies that despite the government's efforts to enforce racial equality, segregation persists. The cartoon likely aims to highlight the ineffectiveness of such policies and the ongoing racial disparities in society. Therefore, the most suitable deep meaning is that the cartoon satirizes that the I.R. law is a road to nowhere, yet the government insists on enacting and enforcing it.

Final answer: A \textcolor{red}{\ding{55}} 
&
\small\raggedright
The image depicts a train with separate compartments labeled \"CAR FOR WHITE FOLKS\" and \"JIM CROW CAR,\" suggesting segregation and discrimination based on race. This imagery conveys a critique of racial inequality and the systemic racism embedded in societal structures. The cartoon highlights the harsh realities of racial segregation, reflecting the deep-seated issues of discrimination and the loss of individuality due to societal norms. Option A focuses on the I.R. law, which doesn't directly relate to the cartoon's theme of racial discrimination. Option B emphasizes societal change, but it doesn't capture the specific critique of racial discrimination depicted in the cartoon. Option C discusses the loss of individuality in educational systems, which is not directly relevant to the cartoon's focus on racial segregation. Option D accurately encapsulates the cartoon's critique of racism and segregation, aligning perfectly with the image's message. Therefore, option D is the most suitable choice as it directly addresses the cartoon's deep meaning of racism.

Final answer: D \textcolor{green}{\ding{51}} 

\end{tabular}
\end{table*}


\begin{table*}
\centering
\caption{Qualitative comparison on cartoon question 5.}
\label{apppic5}
\setlength{\tabcolsep}{4pt} 
\begin{tabular}{@{}m{0.45\linewidth} m{0.45\linewidth}@{}}
\toprule
\multicolumn{1}{c}{\textbf{Image}} & \multicolumn{1}{c}{\textbf{Question}} \\
\midrule
\includegraphics[width=\linewidth]{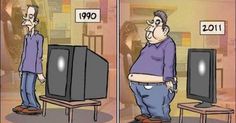} &
\small
A. The moral of this cartoon is that people’s physical health is affected by their daily eating habits; on a deeper level, people’s thoughts are also related to what people usually read, and what we read and come into contact with will affect us in turn.

B. The comic strip implicitly raises concerns about the impact of technological advancements on human health and lifestyle. We should change relationship between technology and human bodies, highlighting the importance of prioritizing physical health and balance in an increasingly digital age.

C. Electronic devices like TV confine humanity. We should get rid of electronic devices and embrace nature.

D. This cartoon satirizes many people in society who hope for change, but every one of them refuses to change themselves.
\\
\toprule
\multicolumn{1}{c}{\textbf{Qwen2.5VL-3B}} & \multicolumn{1}{c}{\textbf{Ours }} \\
\midrule
\small\raggedright
The image shows a cartoon of a man in 1990 standing next to a small television, while in 2011, the same man is much larger and overweight, standing next to a much larger television. This visual progression suggests a correlation between the size of the television and the man's weight gain over time. The moral of the cartoon can be interpreted as a commentary on the impact of technology, particularly television, on human health and lifestyle. The larger television in 2011 could symbolize the increased consumption of media and the sedentary lifestyle that often accompanies it, leading to weight gain. Therefore, the most suitable deep meaning for the image is that people’s physical health is affected by their daily eating habits; on a deeper level, people’s thoughts are also related to what people usually read, and what we read and come into contact with will affect us in turn.

Final answer: A  \textcolor{red}{\ding{55}} 
&
\small\raggedright
The image shows a person in 1990 standing next to a small television, while in 2011, the same person is much larger and overweight, standing next to a much larger television. This visual metaphor suggests a correlation between increased screen time and physical health issues. The cartoon highlights how technological advancements have influenced human behavior and health. The deeper meaning touches on the impact of daily habits and the relationship between technology and human health. Option A focuses on the impact of eating habits, which aligns with the cartoon's theme but doesn't fully capture the broader technological impact depicted. Option B emphasizes the impact of technology on health and lifestyle, which resonates well with the cartoon's message.Option C suggests a rejection of electronic devices, which misses the cartoon's critique of technology.Option D addresses societal change but doesn't directly relate to the cartoon's core message about technology and health. Therefore, option B is the most suitable choice as it encapsulates the cartoon's deeper meaning about technology's influence on health.

Final answer: B \textcolor{green}{\ding{51}} 
\end{tabular}
\end{table*}


\begin{table*}

\centering
\caption{Qualitative comparison on cartoon question 6.}
\label{apppic6}
\setlength{\tabcolsep}{4pt} 
\begin{tabular}{@{}m{0.45\linewidth} m{0.45\linewidth}@{}}
\toprule
\multicolumn{1}{c}{\textbf{Image}} & \multicolumn{1}{c}{\textbf{Question}} \\
\midrule
\includegraphics[width=\linewidth]{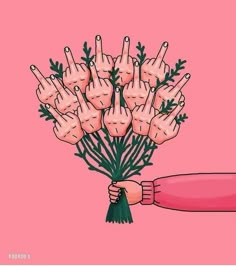} &
\small
A. This picture satirizes some people who talk too much when they give a speech, but what they actually say is illogical and ridiculous. 

B. If the person who sends flowers has ill intentions, then the flowers they send will only be offensive to people. No matter how well they are packaged, ill will is always hurtful.

C. With flowers in your heart, your mood will naturally get better. We should be able to adjust our mood.

D. This picture uses a sarcastic attitude to express the problem of people nowadays who only talk but don't take action. Without people who actually do things, all the talk from the critics and talkers is meaningless.
\\
\toprule
\multicolumn{1}{c}{\textbf{Qwen2.5VL-3B}} & \multicolumn{1}{c}{\textbf{Ours }} \\
\midrule
\small\raggedright
The image shows a hand holding a bouquet of flowers with middle fingers pointing upwards. This visual metaphor is often used to convey a message of defiance or rebellion. Let's analyze each option in relation to this image: A. This picture satirizes some people who talk too much when they give a speech, but what they actually say is illogical and ridiculous. This option suggests that the image is a critique of people who speak excessively but have no substance. However, the image does not seem to focus on the content of the speech but rather on the gesture of the middle fingers. B. If the person who sends flowers has ill intentions, then the flowers they send will only be offensive to people. No matter how well they are packaged, ill will is always hurtful. This option implies that the flowers themselves are offensive due to the gesture of the middle fingers. However, the flowers are not the focus of the image; the gesture is. C. With flowers in your heart, your mood will naturally get better. We should be able to adjust our mood. This option suggests a positive message about adjusting one's mood with flowers. However, the image does not convey this message; it uses the gesture of the middle fingers to make a point. D. This picture uses a sarcastic attitude to express the problem of people nowadays who only talk but don't take action. Without people who actually do things, all the talk from the critics and talkers is meaningless. This option aligns with the image's message. The gesture of the middle fingers, often associated with rebellion or defiance, can be interpreted as a critique of people who talk a lot but do little. The flowers, while not the main focus, can be seen as a symbol of the speaker's intentions, which are not well-received. 

Final answer: D \textcolor{red}{\ding{55}} 
&
\small\raggedright
The image depicts a bouquet of flowers with middle fingers pointing upwards, symbolizing a gesture of defiance or rebellion. This visual conveys a sense of rebellion against conventional norms, suggesting a critique of superficial gestures. The flowers, typically associated with love and affection, are juxtaposed with the aggressive gesture, highlighting the contrast between outward expressions and underlying intentions. The deep meaning of the image likely reflects the critique of superficial acts of communication and the underlying ill intentions behind such gestures. Option A focuses on the critique of speech, but it doesn't directly relate to the aggressive gesture in the image. Option B emphasizes the offensive nature of the gesture, aligning well with the image's message. Option C suggests a positive outlook on flowers, which doesn't resonate with the aggressive symbolism. Option D addresses the problem of talk without action, but it misses the specific critique of superficial gestures depicted in the image. Therefore, option B is the most suitable choice as it encapsulates the image's critique of ill intentions behind gestures.

Final answer: B \textcolor{green}{\ding{51}} 

\end{tabular}
\end{table*}



\subsection{Case Study on MoCoT}
\label{app:case}

\begin{figure*}
  \centering
  \includegraphics[width=1\linewidth]{pic/case2.pdf}
  \caption{Case study comparing standard CoT (Qwen-2.5-3B-VL) and our MoCoT on a visual satire task.}
  \label{case}
\end{figure*}

To illustrate the difference between standard CoT and MoCoT, we analyze the cartoon in Figure~\ref{case}, which contrasts a person using a computer in 1980 with another in 2013. The humor does not arise from any single visual fact alone, but from a context-sensitive comparison across multiple cues: the devices become thinner and more advanced, while the user’s posture, body shape, and lifestyle implications suggest increasing sedentary behavior and declining physical well-being. The intended meaning is therefore satirical rather than literal, criticizing the hidden human cost behind technological convenience.

Standard CoT tends to compress this interpretation process into a single answer-oriented reasoning chain too early. It can identify locally plausible cues such as device modernization, improved comfort, or environmental implications, but fails to preserve and jointly resolve the intermediate interpretive judgments required for the cartoon’s satire. As a result, it produces a rationale that remains superficially reasonable yet misses the core cross-cue incongruity between technological refinement and human physical degradation.

In contrast, MoCoT constructs the rationale through a staged process that is more faithful to the structure of comic understanding. It first decomposes the original question into complementary sub-questions that make explicit several intermediate interpretive needs, such as changes in posture, body shape, product design, and their implied lifestyle consequences. It then answers each sub-question through localized image-grounded reasoning, so that intermediate conclusions are tied to concrete visual evidence rather than immediate answer speculation. Instead of directly aggregating these sub-results, MoCoT further performs reflective meta-reasoning to assess which sub-answers are reliable and how they should be integrated into a coherent global interpretation. This process allows the model to infer the central satirical message: although technology becomes more streamlined and convenient, human users become more sedentary and physically degraded. Finally, post-hoc verification helps ensure consistency between the final rationale and the selected answer. This example shows that MoCoT better preserves the multi-cue and context-sensitive nature of comic reasoning, enabling more faithful interpretation of visual satire than standard CoT.

\end{document}